%% file: acl_latex.tex
\pdfoutput=1

\documentclass[11pt]{article}

\usepackage[final]{acl}

\usepackage{amsfonts}
\usepackage{pifont}
\usepackage{listings}
\usepackage{amsmath}
\usepackage{times}
\usepackage{latexsym}
\usepackage{multirow}
\usepackage{verbatim}

\usepackage[T1]{fontenc}

\usepackage[utf8]{inputenc}

\usepackage{microtype}

\usepackage{inconsolata}

\usepackage{graphicx}

\title{
CoachMe: Decoding Sport Elements with a Reference-Based Coaching Instruction Generation Model}

\usepackage{algorithm}
\usepackage{algorithmicx}
\usepackage{algpseudocode}

\author{
  \textbf{Wei-Hsin Yeh\textsuperscript{1,3}},
  \textbf{Yu-An Su\textsuperscript{1}},
  \textbf{Chih-Ning Chen\textsuperscript{1}},
  \textbf{Yi-Hsueh Lin\textsuperscript{1,2}},
\\
  \textbf{Calvin Ku\textsuperscript{2}},
  \textbf{Wen-Hsin Chiu\textsuperscript{2}},
  \textbf{Min-Chun Hu\textsuperscript{2}},
  \textbf{Lun-Wei Ku \textsuperscript{1}},
\\
  \textsuperscript{1}Institute of Information Science, Academia Sinica,
\\
  \textsuperscript{2}National Tsing Hua University,
  \textsuperscript{3}National Taiwan University
\\
    \texttt{\{weihsinyeh168, allen0512911\}@gmail.com},
    \\
    \texttt{\{yuansu, andrewman71, lwku\}@iis.sinica.edu.tw},
    \\
    \texttt{calvinku@gapp.nthu.edu.tw}, 
    \texttt{whchiu@mx.nthu.edu.tw},
    \texttt{anitahu@cs.nthu.edu.tw} 
}

\begin{document}
\maketitle

\input{latex/A_Abstraction/abstract}

\input{latex/B_Introduction/Intro}
\input{latex/C_Related_work/Related_work}

\input{latex/D_Methodology/Methodology}

\input{latex/E_Experiments/Experiments}

\input{latex/F_Results/Results}


\input{latex/H_limitation/limitation}

\bibliography{acl_latex}

\input{latex/I_Appendix/appendix}

\end{document}

%% file: latex/A_Abstraction/abstract.tex
\begin{figure*}
\centering
\includegraphics[width=1\textwidth]{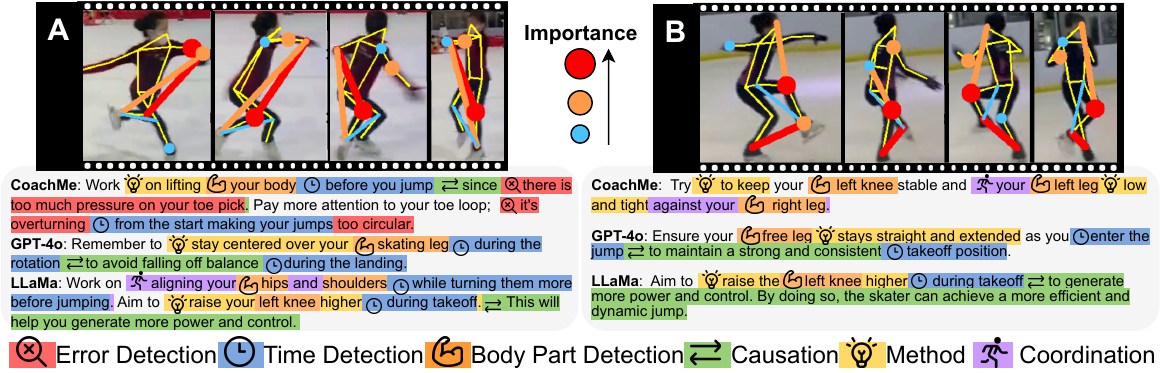} 
\caption{Each video is represented as a sequence of temporally ordered
images, with a visualized attention graph of Human Pose Perception (see~\ref{Human Pose Perception}) overlay. Accompanying each sequence
are instructions generated by three models---CoachMe, LLaMa, and 
GPT-4o---annotated with pictograms that highlight evaluation indicators assessing sport
utility and semantic relevance.}
\label{fig:Demo}
\end{figure*}

\maketitle
\begin{abstract}
Motion instruction is a crucial task that helps athletes refine their technique by analyzing movements and providing corrective guidance. Although recent advances in multimodal models have improved motion understanding,
generating precise and sport-specific instruction remains challenging due to the highly domain-specific nature of sports and the need for informative guidance. We propose CoachMe, a reference-based model that analyzes the differences between a learner’s motion and a reference under temporal and physical aspects. This approach enables both domain-knowledge learning and the acquisition of a coach-like thinking process that identifies movement errors effectively and provides feedback to explain how to improve. In this paper, we
illustrate how CoachMe adapts well to specific sports such as skating and boxing by learning from general movements and then leveraging limited data. Experiments show that CoachMe provides high-quality instructions instead of directions merely in the tone of a coach but without critical information. CoachMe outperforms GPT-4o by 31.6\% in G-Eval on figure skating and by 58.3\% on boxing. Analysis
further confirms that it elaborates on errors and their corresponding improvement methods in the generated instructions. You can find CoachMe here: \url{https://motionxperts.github.io/}
\end{abstract}

%% file: latex/B_Introduction/Intro.tex
\section{Introduction}
Given their strong ability to connect vision with language, recent multimodal
models for motion-related tasks have shown significant progress.
Existing efforts primarily focus on motion caption~\cite{9561519,
zhang2023generating} or universal models that perform any
motion-related tasks~\cite{guo2022tm2t, jiang2024motiongpt, li2024unimotion}.
These models, trained on large, high-quality datasets such as 
HumanML3D~\cite{Guo_2022_CVPR} and KIT-ML~\cite{plappert2016kit}, excel at understanding
motion and generating descriptions such as ``A man lifts his left knee to his
right elbow.''
This capability is important for tasks that require subtle movement analysis
such as motion tracking~\cite{karaev24cotracker3}, physics-based character
control~\cite{10.1145/3618404} in animation, and robotics. Sports is another
area where it is essential to recognize how specific gestures affect overall motion.
Excellence in performance relies on movements executed with high
precision and outstanding coordination, encompassing both temporal and
positional accuracy. Although expert coaches can provide the effective
instructions that athletes covet, they are not always readily available.

One solution in scenarios where coaching resources are limited is automatic generation 
of precise motion instructions. However, there are two
challenges in this task: The first is the dynamics of sports, where
each discipline has unique movement patterns. Experts gain domain knowledge
through years of studying professional 
techniques~\cite{10.1109/TVCG.2022.3230855, 10.1007/s11042-023-14490-2}. To replicate this
expertise, a model must analyze poses from multiple perspectives across
different sports, such as changing joint angles, orientations, and temporal
variations within a single motion. Note that a motion's meaning differs
across sports: in skating, the coordination of knee and shoulder ensures
balance and jump execution, whereas in boxing, force transfer from foot to fist
determines punching power and strategy. 

The second challenge is providing highly informative instructions. Coaches
leverage years of experience to guide movement adjustments at precise moments.
To achieve similar effectiveness, a model must analyze both physical and
temporal perspectives and provide sport-specific, actionable feedback. This
includes identifying incorrect body parts, the degree of adjustment, and
refining joint alignment or time error, as shown in Fig.~\ref{fig:Demo}.

In response to these challenges, we propose CoachMe, a model that obtains
domain-specific knowledge from limited data by comparing a learner's motion to
a reference motion. By analyzing motion differences and leveraging its
intrinsic understanding of movement, CoachMe identifies opportunities for
improvement. Furthermore, Basic CoachMe, the base model, facilitates 
adaptation to sports such as skating and boxing.
Moreover, CoachMe's workflow emulates the structured reasoning of a professional coach while instructing athletes, and hence addresses the second
challenge. This workflow enables CoachMe to integrate both temporal and
physical information, allowing it to generate instructions that not only
identify the timing of mistakes but also explain how to improve.

We visualize the weights learned by the model between joints and moving
directions and generate instructions to determine whether CoachMe understands the
issues in executing the current movement. Additionally, we conduct quantitative
and qualitative experiments as well as human evaluation of the performance of
CoachMe. Our contributions are threefold: (1)~We propose CoachMe, the first
reference based model that learns motion differences 
and generates instructions automatically to offer great precision and sport
utility; (2)~We conduct extensive experiments to validate the generated
instructions by comparing CoachMe to state-of-the-art vision language models, including human evaluation by experts, which is typically challenging to obtain; 
(3)~We construct datasets for instruction generation on two sports---figure skating
and boxing---including videos, instructions, and labeled error segments by
professional coaches. Datasets will be available upon acceptance.


%% file: latex/C_Related_work/Related_work.tex
\section{Related Work}

\begin{figure*}[ht]
\centering
\includegraphics[width=1\textwidth]{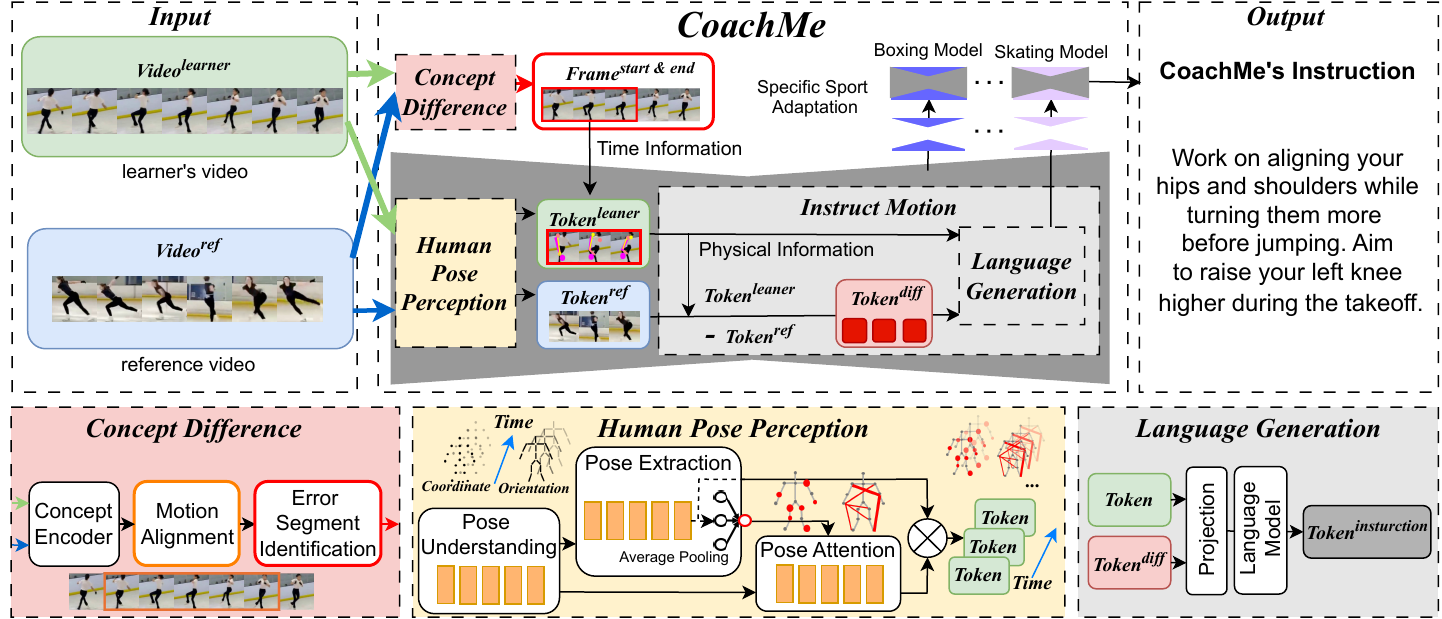} 
\caption{CoachMe architecture comprises three modules: Concept 
Difference (Sec.~\ref{sc:ss}), Human Pose Perception (Sec.~\ref{sc:human pose perception}), and
Instruct Motion (Sec.~\ref{sc:Instruction Generation}). Instruct Motion compares
the motion $\mathit{Token}^{\mathit{learner}}$ with $\mathit{Token}^{\mathit{ref}}$ to obtain the difference
$\mathit{Token}^{\mathit{diff}}$ and take $\mathit{Token}^{\mathit{learner}}$
and $\mathit{Token}^{\mathit{diff}}$ as input to the LM to generate
instructions.}
\label{fig:framework}
\end{figure*}

Vision language models~\cite{liu2023llava, zhang2023llamaadapter,
2023videochat, damonlpsg2023videollama, lin2023video} have demonstrated strong
capabilities in video-to-text tasks as well as their inverse: text-to-video
generation. However, these approaches generally overlook the unique
characteristics of pose features and fail to consider the specific demands of
different sports. For instance, in skating, certain aspects require special
attention, yet existing models tend to generate instructions that are overly
general and broadly applicable rather than tailored to the nuances of
individual activities.

Beyond general video analysis, human motion modeling has gained traction
as a crucial aspect of video understanding. Studies on 
text-to-motion~\cite{petrovich23tmr, TEACH:3DV:2022, chen2023executing} generate 3D motion
from language descriptions, whereas motion-to-text 
approaches~\cite{zhang2023generating, li2024unimotion} describe motion in natural
language. For instance, TM2T~\cite{guo2022tm2t} and 
MotionGPT~\cite{jiang2024motiongpt,wang2024motiongpt2generalpurposemotionlanguagemodel} use vector quantized-variational 
autoencoders~\cite{DBLP:journals/corr/abs-1711-00937} to map motion sequences to discrete
representations for motion-related tasks. However, these models focus on
general motion descriptions rather than detailed coaching instructions. 
\citet{liu2022posecoach} and 
\citet{tanaka2023automaticedgeerrorjudgment} refine specialized
skills but are limited to specific sports and rely primarily on visual feedback
techniques rather than textual feedback. CoachMe overcomes these limitations by
integrating textual instructions with visual explanations, as shown in
Fig.~\ref{fig:Demo}, while also developing a workflow that is adaptable to different
sports.

The idea of concepts is often used for challenging tasks. 
Concept-based techniques apply single concepts to analyze humor or
identify patterns through activation
functions~\cite{tennenholtz2024demystifying, kim2018interpretability} to
achieve promising results. In this paper, we extend this to two concepts---\textit{motion}
and \textit{difference}---in the sport technology domain. CoachMe aligns
videos~\cite{dwibedi2019temporal,chen2022framewise,kwon2022contextaware} to
facilitate synchronization. During this process, motion concept embeddings and
difference concept embeddings are generated for further analysis of movement discrepancies.

%% file: latex/D_Methodology/Methodology.tex
\section{Methodology} \label{sc:method}
To construct the thought process of a coach guiding an athlete, we propose 
generating instructions based on the differences between the
input video and the reference video. CoachMe consists of three modules:
Concept Difference, which generates the difference concept; Human Pose
Perception, which extracts the motion concept; and Instruct Motion, which 
processes and integrates the two concepts, as shown in 
Fig.~\ref{fig:framework}.
\input{latex/D_Methodology/Concept_Difference}

\input{latex/D_Methodology/Human_pose_perception}

%% file: latex/D_Methodology/Concept_Difference.tex
\subsection{Concept Difference Module}
\label{sc:ss}
\label{sc:mdbv}

\paragraph{Concept Encoder}
We define a \textit{concept} as a quantification of performed action, regardless of their viewpoint. A \textit{concept difference} is the deviation between two performances. The
concept difference between two frames is computed as
\begin{equation}
  \label{eq:concept_generation}
  {c = \mathrm{F}(x_{r}) - \mathrm{F}(x_{l}) },
\end{equation}
where $x_{l}$ represents a frame from the learner’s clip, $x_{r}$ is
the corresponding reference frame, and $\mathrm{F}$ is a Concept Encoder, which we adopted CARL \cite{chen2022framewise}. Concept Encoder was trained independently and was frozen during instruction generation training.

\paragraph{Motion Alignment}
Motion Alignment identifies the interval in the learner video $V^L$ that best corresponds to the reference video $V^R$ and the Concept Difference embedding $C$ is then obtained by subtracting the aligned segment from the reference segment, as shown in Algorithm \ref{algo:motion_alignment}.

\begin{algorithm}
\caption{Motion Alignment}\label{algo:motion_alignment}
\begin{algorithmic}
\Procedure{$DTW$}{$A$,$B$,$d_a$,$d_b$}
\State $distance, j \gets \infty,  -1$
\State {\textbf{For} $i \in \{0 , 1 , ..., {d_a - d_b}\}$ \textbf{do:}}
\State {\ \ $tmp \gets$D$(A_{i:i+d_b},B_{0:d_b})$} \Comment{D:cost matrix}
\State {\ \ \textbf{If} $tmp < distance$  \textbf{:}}
\State {\ \ \ \ \ \ \ $distance, j \gets tmp, i$}
\State \textbf{return} $j$ \Comment{Optimal interval's start frame}
\EndProcedure
\State $\textbf{Input}:V^L, V^R$
\State $d^L, d^R \gets length(V^L), length(V^R)$
\State {\textbf{If} $d^L < d^R$ \textbf{:}}
\State {\ \ \ $V^L, V^R\gets V^R,V^L$}
\State {\ \ \ $d^L, d^R \gets length(V^L), length(V^R)$ }

\State $C^L,C^R\gets $F$(V^L),$F$(V^R)$\Comment{F:Concept Encoder}
\State $j \gets DTW(C^L,C^R,d^L,d^R)$
\State $C^{L'} \gets C^{L}_{j:j+d_R}$
\State $c_{t} \gets C^R_t - C^{L'}_t , t\in\{0,1...d_R\}$
\State $C \gets c_{t}, t\in\{0,1...d_R\}$
\State \textbf{return} $C$\Comment{Concept Difference}
\end{algorithmic}
\end{algorithm}



\paragraph{Error Segment Identification}
This module identifies error-prone segments within the target motion. Since
frames with higher concept differences are more likely to need correction, we
train a model using Concept Difference embeddings $C$ to predict error
segments.
Given the temporal dependencies in motion sequences, we implement the error
segment selection model using a transformer encoder, which captures frame-wise
features and temporal relationships. The model outputs the interval requiring
instruction as
\begin{equation}
\label{eq:error_segment}
R_{i}' = R_{i_{\mathit{SOE:EOE}}},
\end{equation}
where SOE and EOE denote the start and end of the identified error segment.

%% file: latex/D_Methodology/Human_pose_perception.tex
\subsection{Human Pose Perception Module} 
\label{sc:human pose perception}
Human Pose Perception module contains 3 submodules: Pose Understanding $\mathrm{PU}$, Pose
Extraction $\mathrm{PE}$, and Pose Attention $\mathrm{PA}$. Each submodule is based on
graph convolutional networks. We first clip $V^L$ according to $R_{i}'$, which is the temporal information from Concept Difference. We employ HybrIK~\cite{li2022hybrik} predicts 22 joint coordinates ${J}$. We chose HybrIK as it gives better performance when used as a pose estimator of CoachMe. Its inference time is also acceptable since users receive results asynchronously. Details will be discussed (Sec. \ref{PoseEstimator}).
Next, subtracting $J$
outside from $J$ inside results in the joint orientations, $O$:
\begin{equation}
    O_{a,b}= J_a - J_b \quad a,b \in \{0,1, \ldots, 21\}
\end{equation}

To capture the mutual influence of $J$ and $O$, $\mathrm{PU}$ shares weights during training. Additionally, to interpret the physical information
of motion, $\mathrm{PU}$ is trained on ${G_S}$, which represents the graph layout of the
human skeleton. ${G_S}$ is constructed from adjacent joint pairs and their
distances. Inspired by STA-GCN~\cite{Shiraki_2020_ACCV}, this approach
incorporates both the temporal dynamics of the motion sequence and the spatial
relationships within the skeletal structure. Consequently, $\mathrm{PU}$ learns
representations of $J$ and $O$ and transforms them into the motion token $T$.
$\mathrm{PE}$ is also trained on ${G_S}$ and generates the motion token $T'$, which
mainly considers local body parts:
\begin{equation}
   T' = \mathrm{PE}_{G_S} (T);\ T = \mathrm{PU}_{G_S} (O \oplus J),
\end{equation}
where $\oplus$ denotes concatenation.
$\mathrm{PE}$ applies average pooling on $T'$ to generate the attention joint ${J_A}$
and attention graph ${G_A}$ which contain latent relationships between the
originally disconnected joints and identifies key joints and
relationships. The visualized attention graph, which consists of $J_A$ and
$G_A$, is provided in Figure~\ref{fig:Demo}.
\begin{equation}
    G_A, J_A = \mathrm{Pool}_{\mathit{avg}} (T')
\end{equation}
In $\mathrm{PA}$, we use $J_A$ as the input for $\mathrm{PA}_{G_A}$ and train it solely with
$G_A$, ensuring a focus on key relationships extracted from $G_A$. Unlike
STA-GCN, CoachMe trains $\mathrm{PA}$ solely on $G_A$. Since $\mathrm{PE}$ has already been
trained with $G_S$, it sufficiently captures the physical information within
local body parts. This allow $\mathrm{PA}$ to focus on learning the physical dynamics of
the global body structure. With ${J_A}$ and ${G_A}$, $\mathrm{PA}_{G_A}$ propagates
the essential attributes of key joints across all previously disconnected
joints and local body parts:
\begin{equation}
    T'' = \mathrm{PA}_{G_A} (J_A \cdot T),
\end{equation}
where $\cdot$ denotes the dot product.
We utilize both local motion tokens $T'$ and global motion tokens $T''$ as the
final motion tokens, incorporating local, global, spatial, and temporal
information:
\begin{equation}
    \label{eq:concate}
    \mathit{Token} = T' \oplus T''.
\end{equation}
Consequently, Human Pose Perception captures precise motion actions and analyzes
subtle motion variations which influence body coordination and are helpful
in detecting 
  poor posture.  

\subsection{Instruct Motion Module}

\label{sc:Instruction Generation}
\begin{figure}[t]
\centering
\includegraphics[width=1\columnwidth]{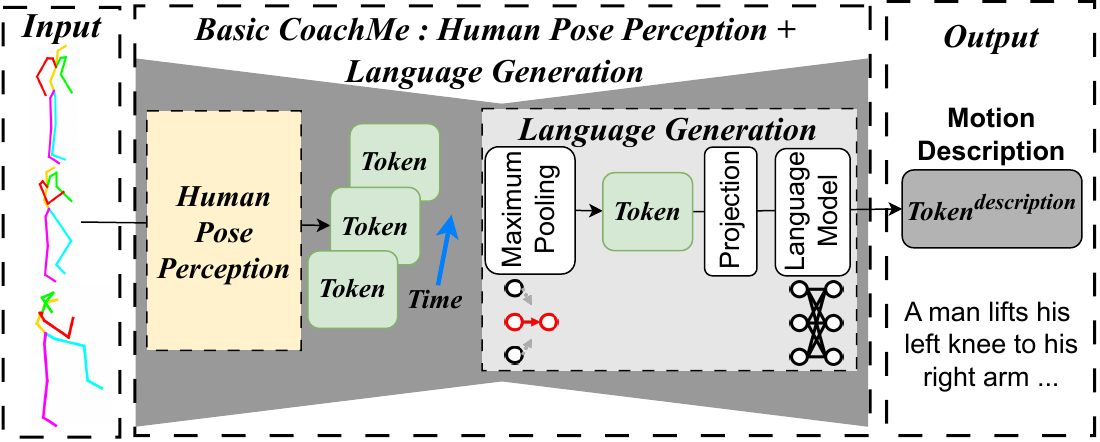} 
\caption{Basic CoachMe, which consists of the
Human Pose Perception and Language Generation modules, 
performs tasks related to motion description.}
\label{fig:Generation}
\end{figure}
Basic CoachMe was initially pretrained on the HumanML3D dataset to understand
the motion token $\mathit{Token}$ and to learn how to generate textual descriptions of
motion, as shown in Fig.~\ref{fig:Generation}. To enable CoachMe to provide
motion-specific instructions, we integrate temporal segment information
from Concept Difference and physical information from Human Pose Perception.
We thus obtain the concept motion token $\mathit{Token}$,
which represents motion information within the error interval. Next, we compute
the concept difference token $\mathit{Token}^{\mathit{diff}}$ by subtracting the learner's
motion token from the reference's motion token. 

$\mathit{Token}$ and $\mathit{Token}^{\mathit{diff}}$ first undergo maximum pooling along the
temporal dimension,which captures each motion token at the
most critical moment of the motion sequence that requires further refinement.
Next, they are projected into the same latent space. In Projection, understanding the relationship between motion and differences 
helps identify which aspects of motion need improvement. To enhance adaptability, when fine-tuning the pretrained weights, Basic CoachMe, for the motion instruction task, we
apply low-rank adaptation (LoRA)~\cite{DBLP:journals/corr/abs-2106-09685} to Human Pose Perception and Projection. This aims to refine Human Pose Perception to
recognize sport-specific motion patterns and adjust the Projection in
Instruct Motion to interpret the relationship of $\mathit{Token}^{\mathit{diff}}$ and $\mathit{Token}$.
Finally, the two types of tokens are transformed into instruction~$I$
through a language model $\mathrm{LM}$. Because our sports dataset is relatively 
small~\cite{brigato2020closelookdeeplearning}, we chose the low-complexity 
T5~\cite{DBLP:journals/corr/abs-1910-10683}, which has only 223M parameters, as
our language model $\mathrm{LM}$:
\begin{equation}
    I = \mathrm{LM}(\mathrm{Proj}(\mathrm{Pool}_{\mathit{max}}(\mathit{Token} \oplus {\mathit{Token}^{\mathit{diff}}})),
\end{equation}
where $\mathrm{Proj}$ denotes the projection layer.

%% file: latex/E_Experiments/Experiments.tex
\section{Experiments}

\input{latex/E_Experiments/datasets}
\input{latex/E_Experiments/settings}

\input{latex/E_Experiments/experimental_design}

%% file: latex/E_Experiments/datasets.tex
\subsection{Datasets} 
\label{sc:datasets}
Three datasets were utilized in this paper. \textbf{HumanML3D} was adopted for
learning general movements, whereas the \textbf{Figure Skating} (FS) and \textbf{Boxing} (BX) datasets,
which we collected, were specifically created for learning sport-specific
elements. HumanML3D~\cite{Guo_2022_CVPR} contains motion sequences from
HumanAct12~\cite{guo2020action2motion} and AMASS~\cite{mahmood2019amass}. 
FS comprises 4 types of skating jump videos: single
Axel, double Axel, Lutz, and Loop from single learners, annotated by single figure skating coach. Each video is labeled with instructions and corresponding intervals
to identify errors, which are denoted as ground truth clips (GT clips), as shown in Table \ref{tab:datasets}. Ground truth clips are employed as the learning targets for the Error Segment Identification module.
BX comprises 2 types of boxing technique video: Jab and Cross from multiple learners, annotated by 3 boxing coaches. Each video is labeled with instructions without an error segment. To enhance diversity, we employed GPT-4o to augment the instructions annotated by coaches, as detailed in Section~\ref{Appendix:Data Augmentation Template}. A summary of these datasets is presented in Table~\ref{tab:datasets}.
Reference videos were sourced from YouTube coaching content and validated by professional coach. 


\begin{table}[t]
\centering
\small
\begin{tabular}{c| c c c c}
\hline
\textbf{Dataset} & \# video & \# of & GT & Aug.\\
 &train : test&motion&inst.&{}\\ \hline
HumanML3D & 23384:4384 & - &  avg. 3 & no \\
BX & 163 : 41 & 2 & 3 & 3 \\
FS & 177 : 40 & 4 & 1 & 0 \\
FS(GT clips) & 449 : 64 & 4 & 1 & 5 \\
\hline
\end{tabular}
\caption{Datasets used in experiments. The
number of training and test samples are presented for each dataset. FS(GT clips) denotes as ground truth clips that error segments annotated by coach. GT inst. denotes as the number of ground truth instructions. Aug. denoted as the number of augmented instructions for one video.}
\label{tab:datasets}
\end{table}


%% file: latex/E_Experiments/settings.tex
\subsection{Settings}
We pretrained Basic CoachMe on HumanML3D to accomplish the motion description task and then adapt it to the FS and BX datasets for motion instruction. To ensure a consistent coordinate system across 3 datasets, we set the pelvis joint as the origin and adopt a local coordinate system. To assess whether referencing also aids descriptions, we experimented with 2 strategies on motion description task:
(1)~using the first frame of each motion sequence as its reference.
(2)~employing zero padding to allow the adapted model to learn reference information independently.

We further investigate how the \textbf{$Token^{diff}$}, a key component for contrasting correct and incorrect performances, behaves across modalities. This analysis is motivated by the design of CoachMe, which integrates two distinct modalities: RGB for the Concept Difference and skeleton-based features for the Human Pose Perception. We define $2$ settings: CoachMe, which computes $Token^{diff}$ from skeleton-based $Token$ via Human Pose Perception, and CoachMe (RGB), which computes $Token^{diff}$ from RGB-based $Token$ via the Concept Encoder. We compared CoachMe with GPT-4o \cite{hurst2024gpt}, LLaMa 3.2 \cite{dubey2024llama}, ViLa \cite{lin2024vila} and MiniCPM \cite{hu2024minicpm}.

%% file: latex/E_Experiments/experimental_design.tex
\subsection{Experimental Design}
Experiments on description and instruction generation were conducted to
evaluate the model's capability in understanding general movements and
providing coaching for specific sports.
Two experiments were conducted for description generation:
(1)~Comparing CoachMe to SOTA models for motion description generation.
(2)~Assessing using references versus not using them when generating descriptions. 
For the instruction generation task, we conducted four experiments:
(1)~Comparing CoachMe to SOTA models for motion instruction generation.
(2)~Studying the gain from applying concept difference.
(3)~Evaluating how error segment identification (Sec.~\ref{sc:ss}) helps in instruction generation.
(4)~Investigating the impact of different modalities of $Token^{diff}$ on model performance.

\subsection{Indicators for Sport Utility Evaluation}
\label{Disscussion:Instruction Indicators}
\par To assess the sport utility of the generated instructions, we introduce a set of evaluation indicators that can be applied to various sports. 
Based on the findings derived from studies on professional 
athletes~\cite{DBLP:journals/corr/abs-1709-09131, DBLP:journals/corr/abs-1711-09562}, effective instruction must fulfill two key aspects, each comprising three essential indicators: The first aspect is that the instruction must \textbf{clearly define the problem}, ensuring that athletes understand the issue in their movement. This includes:
(1) Detecting \textbf{errors} in the motion; (2) Identifying \textbf{timing} information; (3) Recognizing \textbf{body part} movements.
The second aspect is that the instruction must \textbf{provide a solution}, offering actionable guidance to help athletes correct their movements. This includes:(4) Identifying \textbf{causal relationships} in the sport; (5) Explaining \textbf{how to improve} the sport; (6) Describing how body parts \textbf{coordinate or interact}. A detailed description of these six indicators will be provided in Section \ref{Appendix:Indicator_Detail}.


%% file: latex/F_Results/Results.tex
\begin{table}[t]
    \centering
    \small
    \resizebox{1.0\linewidth}{!}{%
    \begin{tabular}{c|c c c c c c c}
    \hline
      Method & Ref & B1 & B4 & RG  & \textbf{BS} \\ 
      \hline
      TM2T (2022) & x  & {61.7} & \underline{22.3} & \textbf{49.2} & {37.8} \\
      MotionGPT (2023) & x & {48.2} & {12.47} & {37.4}  &  {32.4} \\
      MotionGPT-2 (2024) & x & {48.7} & {13.8} & {37.6}  &  {32.6} \\
      \hline
      STAGCN* & x & \underline{62.8} & {22.1} &{47.0}  & \underline{43.5} \\
      \hline
       {} & x & \textbf{65.4} & \textbf{24.3} & \underline{48.6} & \textbf{45.1} \\   
       {Basic CoachMe} & v &       {62.5} & {20.8} & {43.3}  & {36.8} \\ 
       {} & Pad0 & {59.0} & {18.2} & {41.8}  & {35.6} \\
      \hline
    \end{tabular}
    }
	 \caption{Comparison of motion description on HumanML3D. STAGCN*
	 denotes the combination of STAGCN and Instruct Motion. Basic CoachMe combines Human Pose Perception and Instruct Motion. Ref, B1,
	 B4, RG, and BS denote reference, BLEU-1, BLEU-4, ROUGE, and BertScore,
	 respectively.}
    \label{tab:motion_description_comp}
\end{table}

\begin{table*}[t]
\centering
\small
\begin{tabular}{c|c c c c c c c}
\hline
\textbf{Method} & \textbf{Ref} & \textbf{Error segment} & \textbf{BLEU@1} & \textbf{BLEU@4} & \textbf{Rouge} & \textbf{BertScore} & \textbf{G-eval} \\ 
\hline
\multicolumn{8}{c}{\textbf{Figure Skating (FS)}} \\ \hline
GPT-4o  & x & - & 16.2 & 1.0 & 14.8  & 7.5  & 1.39 \\ 
LLaMa 3.2  & x & - & 12.6 & 0.0 & 11.4  & -4.9  & 1.31 \\ 
MiniCpm &  x & - & 9.7 & 0.0 & 11.4 & 7.2 & 1.37 \\ 
ViLa &  x & - & 8.8 & 1.8 & 11.6 & 11.8  & 1.27 \\ 

\hline
Basic CoachMe & x & Ground truth & 15.0 & 2.5 & 16.9 & 11.0  & 1.53 \\ 

\hline
 & v & Ground truth & \textbf{24.7} & 2.3 & 16.9 & \textbf{26.5} & \underline{1.73} \\ 
CoachMe & v & Aligned Segment & \underline{22.2} & 2.3 & \textbf{20.0} & 11.7 & \textbf{1.83} \\ 
 & v & Error Segment & 20.8 & 2.1 & 15.2 & \underline{20.5} & 1.55 \\ 
\hline

& v & Ground truth & 16.9 & 1.6 & 15.9 & 12.0 & 1.37  \\   
CoachMe (RGB) & v & Aligned Segment & 17.2 & \underline{4.0} & 15.8 & 7.1 & 1.21   \\ 
& v & Error Segment & 19.4 & \textbf{4.4} & \underline{17.0} & 8.0   & 1.57 \\ 
\hline

\multicolumn{8}{c}{\textbf{Boxing (BX)}} \\ \hline
GPT-4o  & x & - & 33.3 & 0.0 & 10.0  & 13.6 & 1.39 \\ 
LLaMa 3.2  & x & - & 16.9 & 0.0 & 11.5  & 3.2 & 1.20 \\ 
MiniCpm & x & - & 9.1 & 1.5 & 9.4 & 9.1 & 1.89 \\
ViLa & x & - & 6.1 & 0.0 & 9.8 & 1.9 & 1.40 \\ 
\hline
Basic CoachMe & x & Aligned Segment & \underline{41.7} & 9.4 & 24.1 & \underline{36.2} & 1.85 \\ 
\hline
\multirow{2}{*}{CoachMe}  & v & Aligned Segment& 38.4 & \underline{12.3} & \textbf{28.4} & 27.2 & \textbf{2.20} \\ 
& v & Error Segment & \textbf{44.5} & \textbf{13.4} & \underline{25.3}  & \textbf{36.9} & 1.61 \\  

\hline
\multirow{2}{*}{CoachMe (RGB)} & v & Aligned Segment & 23.3 & 6.0 & 18.2 & 26.8 & \underline{1.98} \\ 
& v & Error Segment & 13.9 & 0.0 & 11.1 & 16.5 & 1.44 \\ 
\hline
\end{tabular}
\caption{Comparison of motion instruction generation methods on FS and BX.
Ref stands for reference. 
Performance reported across different CoachMe settings and error segment identification
approaches: ground truth (coach-labeled intervals), predicted (model-determined
intervals), and no identification (entire aligned video). We also evaluated
the consistency between the generated instruction and the ground truth using 
G-Eval~\cite{liu2023gevalnlgevaluationusing}.}
\label{tab:combined_results}
\end{table*}

\section{Results}
\subsection{Description Generation}
Table \ref{tab:motion_description_comp} compares the generated descriptions of movements from SOTA methods. Models are all grounded in the same 3D human motions. Basic CoachMe outperforms others in generating movement descriptions. Human Pose Perception captures 3D motions by treating global and local information equally while considering orientation and coordinates. It thus produces more accurate motion representations than STA-GCN. Then, Instruct Motion transforms these motion tokens into motion descriptions. Interestingly, Table \ref{tab:motion_description_comp} shows that motion description performance is not improved by references, suggesting that understanding general movements does not require predefined standards.

\subsection{Sport Instruction Generation}
\label{sec:Sport Instruction Generation}
\par In this section, we evaluate CoachMe on sport-specific instruction generation, using the FS and BX datasets. 
To evaluate the role of \textbf{reference information}, we compare CoachMe's performance with and without the use of reference video. As shown in Table \ref{tab:combined_results}, incorporating reference leads to consistent improvements on both the FS and BX. Human evaluation (see Appendix  \ref{Appendix:Modalities_Human_Evaluation}) supports this finding. This demonstrates the importance of reference-based models in instruction generation.

\par The experimental results presented in Table~\ref{tab:combined_results} indicate that CoachMe consistently achieves higher performance when operating with $Token^{diff}$ based on skeleton modality compared to RGB. The reason is that RGB videos often contain irrelevant elements, such as background distractions, whereas skeleton representations focus solely on motion, leading to more precise instructions for movement refinement. This effect is particularly evident in figure skating, where fast-moving backgrounds introduce additional noise. However, skeleton-based representations may struggle to capture fine-grained joint rotations, such as palm flips.

\par We analyze the impact of the error segment by providing the model with
\textbf{(1) ground truth}: instruction intervals labeled by coaches (available only in the FS dataset, where a figure skating coach provided the ground-truth timestamps, while the boxing coach did not provide such timestamps);
\textbf{(2) aligned segment}: the whole movement is trimmed by the Motion
Alignment module without error segment identification.
\textbf{(3) error segment}: intervals selected by the Error Segment Identification module; 
Note that the Error Segment Identification module achieves an accuracy of $76.14$\%. Take FS as an example, evaluated by comparing correctly predicted frames with those labeled by coaches:
the BertScore results for CoachMe indicate that using either the ground truth or predicted error segments produces instructions that align more closely with
coach-labeled instructions compared to those without error segment identification. This highlights the importance of identifying the most error-prone segment for generating instruction.

\begin{table}
\small
\centering
\begin{tabular}{c c c c}
\hline 
Dataset & dist-1 & dist-2 & dist-3   \\ \hline
\multicolumn{4}{c}\textbf{Ground Truth Instruction} \\ \hline
FS & 0.115 & 0.405 & 0.645 \\ 
BX & 0.037 & 0.115 &  0.182\\ \hline
\multicolumn{4}{c}\textbf{CoachMe's Prediction}  \\ \hline
FS & 0.233 & 0.519 & 0.685 \\ 
BX & 0.070 & 0.136 & 0.172 \\ 
    \hline
    \end{tabular}
    \caption{ Diversity (dist-1, dist-2, dist-3) of ground truth instructions in FS and BX, as well as instructions predicted by CoachMe trained on FS and BX, where dist-n denotes the percentage of distinct n-grams \cite{li-etal-2016-diversity}. Higher distinct scores indicate greater diversity.}
    \label{table:Diversity}
\end{table}

The higher performance observed on BX compared to FS is primarily attributable to the characteristics of the dataset. BX, constructed from $10$ beginner boxers, contains many frequently occurring mistakes, resulting in more consistent feedback from the coaches. Additionally, BX dataset involves only two basic boxing technique with its coaching instructions, leading to lower diversity. In contrast, FS dataset involves $4$ jumping movements, which yields higher instruction diversity, as shown in Table \ref{table:Diversity}, indicating a greater modeling challenge effectively addressed by CoachMe. Consequently, CoachMe naturally inherits the respective diversity from each dataset, as demonstrated by the distinct scores of the generated instructions.

\subsection{Human Evaluation} 
\begin{table}
\small
    \centering
    \begin{tabular}{c c c c}
    \hline 
     Model & Good (\%) & Neutral (\%) & Bad (\%) \\ \hline
     \multicolumn{4}{c}{\textbf{Figure Skating (FS)}} \\ \hline
     GPT-4o & 20.3   &    50.0  &     \textbf{29.7} \\
     LLaMa 3.2 & 18.8      & 45.3  &      35.9 \\
     CoachMe & \textbf{26.6}    &   43.8  &      \textbf{29.7} \\ \hline
     \multicolumn{4}{c}{\textbf{Boxing (BX)}} \\ \hline
     GPT-4o & 46.3   &    17.1  &     36.6 \\
     LLaMa 3.2 & 51.2     & 24.4  &      24.4 \\
     CoachMe & \textbf{56.0}    &   22.0  &      \textbf{22.0} \\
    \hline
    \end{tabular}
    \caption{Comparison of instruction quality for 3 models rated by the human evaluator on the FS and BX datasets.}
    \label{table:win_rate}
\end{table}
\label{app:AOHE}
We work with figure skating coaches and provide professional evaluations. Table~\ref{table:win_rate} demonstrates the results of human evaluation. 
CoachMe received the highest percentage of "Good" ratings (26.6\%), outperforming GPT-4o (20.3\%) and LLaMA (18.8\%).It also aligned more closely with key instructional elements, excelling in identifying the correct timing, relevant body parts, causal relationships, and corrective methods—indicators previously defined in Experiments (Sec~\ref{Disscussion:Instruction Indicators}). These factors consistently appeared when CoachMe was rated as the best, confirming their positive impact. 
Detailed breakdowns can be found in Appendix (Sec.~\ref{FS-human_evaluation-details}). Notably, CoachMe demonstrated the strongest ability to capture coordination-related aspects and analyze complex body mechanics.
\par As for the performance on the BX dataset, CoachMe received the highest percentage of “Good” ratings (56.0\%) and the lowest percentage of “Bad” ratings (22.0\%), as shown in Table~\ref{table:win_rate}.
The superior performance observed on the BX dataset compared to FS is consistent with the findings discussed in Section~\ref{sec:Sport Instruction Generation}.
We interviewed the boxing coach who conducted the human evaluation. The coach emphasized that in sports instruction, concise instructions are more effective and easier for learners to follow.
Moreover, the boxing videos consist of beginners performing $2$ fundamental techniques, making detailed instructions unsuitable.
In contrast, LLaMA and GPT-4o tend to generate overly complex feedback (See \ref{Disscussion:accuracy}.
This preference for concise, targeted instructions likely contributed to CoachMe’s superior performance on the BX dataset. Details can be found (see  Sec.~\ref{BX-human_evaluation-details}).

\section{Discussion}
In this section, we evaluate the generated instructions through indicators (Sec.~\ref{Disscussion:Instruction Indicators}) that assess both their sport utility and semantic relevance. We compare CoachMe to LLaMa and GPT-4o to show its indispensability. Last, we delve deeper into the styles (Sec.~\ref{Disscussion:accuracy}), sport indicators (
Sec.~\ref{Disscussion:factor}) and the design of CoachMe (Sec.~\ref{designOfCoachMe}). 

\subsection{Finding 1: CoachMe Generates More Accurate Instruction}
\label{Disscussion:accuracy}
\begin{figure}
\includegraphics[width=1\columnwidth]{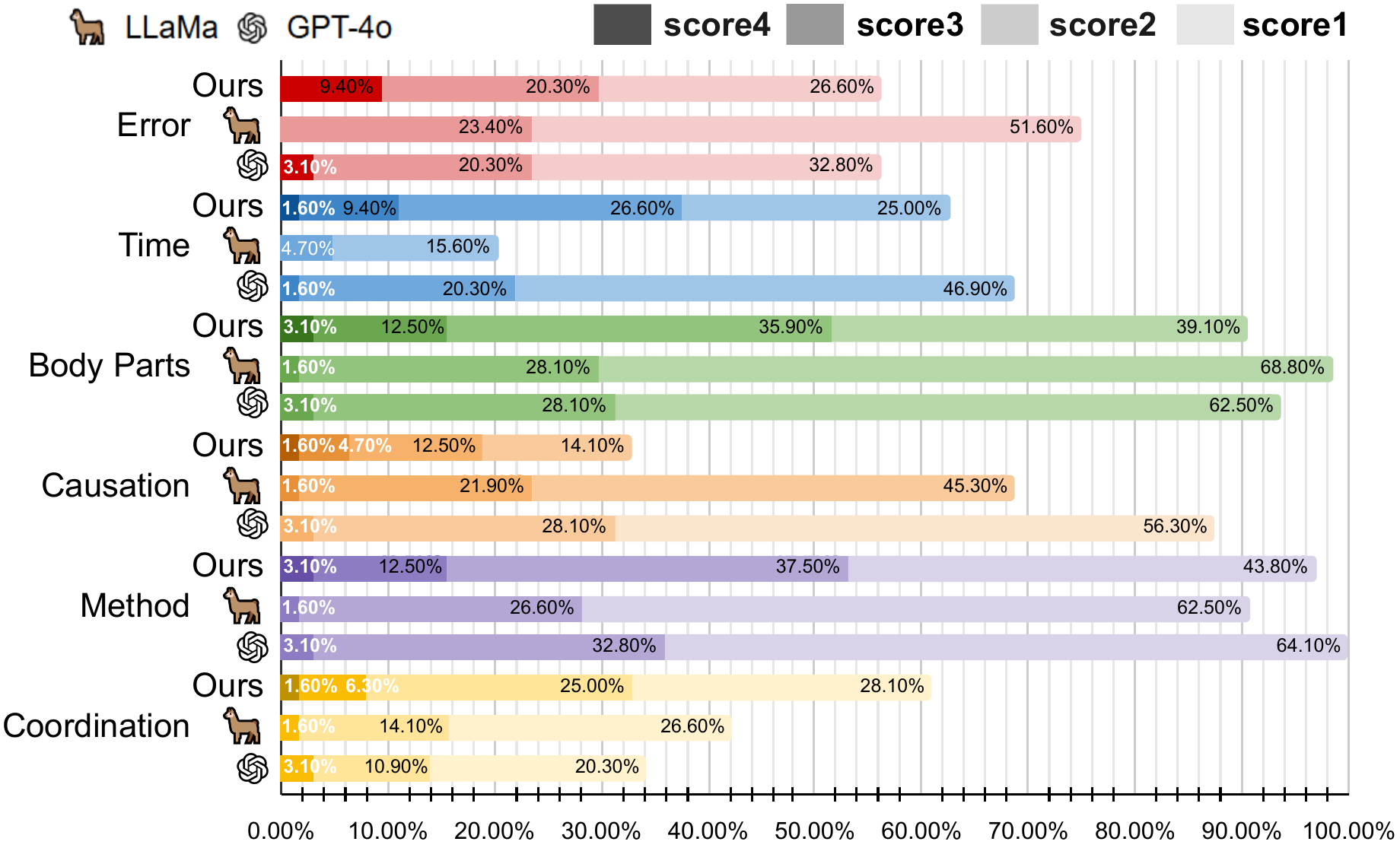}
  \caption {The horizontal bar chart represents the total number of
  instructions assigned to each indicator (Sec.~\ref{Disscussion:Instruction
  Indicators}), with colors indicating their respective G-eval scores. Higher
  G-eval scores indicate more informative and effective guidance that aligns
  more closely with the ground truth. The G-eval scores range from 1 to 5.}
   \label{fig:pie-chart}
 \end{figure}

We analyze the relation between the $6$ indicators and the instruction quality.
GPT-as-a-judge helps annotate the indicators mentioned in the generated
instructions. We compare CoachMe to LLaMa and GPT-4o. 
Figures~\ref{fig:pie-chart} show the percentage of figure skating instructions in which each indicator
is mentioned and the corresponding distribution of their G-eval scores. Details can be found (See Sec. \ref{Appendix:Indicator_Detail}). Boxing indicator frequencies and G-eval score distributions are analyzed in Section~\ref{BX-human_evaluation-details}.
\par The results reveal that although GPT-4o and LLaMa can generate instructions relevant to these indicators (their total bar length is long), they struggle to produce high-scoring instructions (their dark bar length is short). The majority
of their guidance is too general or ineffective, resulting in G-eval scores of 1 or 2 (see their longer light colored bar). In other words, LLaMa and GPT-4o are good at playing the role and instructing in a coach's tone but their content is inaccurate, whereas CoachMe consistently generates precise and actionable instructions. 
Notably, nearly 60.9\% of CoachMe's instructions incorporate coordination-related feedback, highlighting its strength in capturing relationships between body parts. This ability is further reflected in its attention mechanism.

As shown in Fig.~\ref{fig:Demo}, thicker and redder lines indicate more important movement relationships. CoachMe’s attention graph automatically captures joint relations and aligns closely with its instructions. For example,
the model highlights the left leg and right leg connection, which leads to the instruction ``Keep your left leg low and tight,'' demonstrating its ability to attend to relevant body parts for precise instruction.

\begin{figure}[t]
  \includegraphics[width=1\columnwidth]{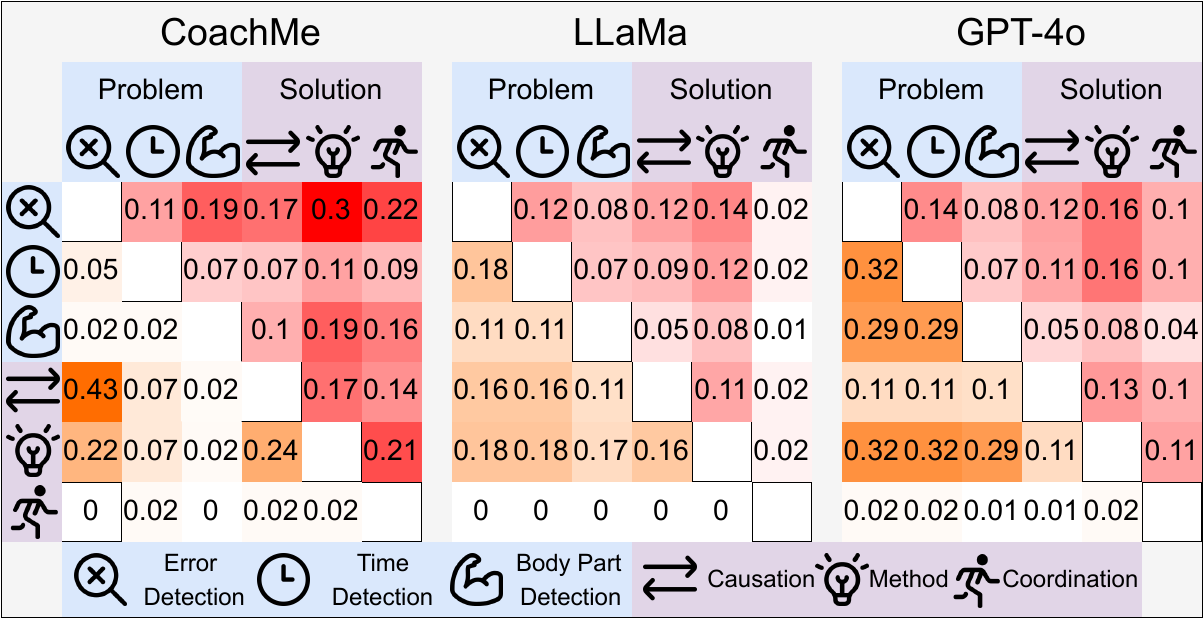}
  \caption {Effectiveness of indicator combinations across models is measured using G-Eval scores, where higher percentages indicate better guidance quality.
  In the evaluation matrix, red numbers in the upper-right triangle represent scores on the FS dataset, highlighting that CoachMe consistently outperforms GPT-4o and LLaMa.
  Orange numbers in the lower-left triangle show scores on the BX dataset.
  Each number denotes the percentage $p$, calculated based on Eq. ~\ref{fig:combination_analyze}.}
   \label{fig:combination_analyze}
 \end{figure}

\subsection{Finding 2: Good Instructions Identify Problems and Provide Solutions}
\label{Disscussion:factor}

As shown in Fig.~\ref{fig:combination_analyze}, we study the pairwise relationship between indicators and investigate which
combinations contribute most to high quality instructions. 
Percentage $p$ in 
Fig.~\ref{fig:combination_analyze} is calculated as
\begin{equation}
    \label{eq:percenrage}
    p = \frac{\Sigma^{N}_{i=1} S\mbox{ \small of instructions containing two indicators}}{\mbox{\small (maximum of $S$}\times N)},
\end{equation}
where $N$ denotes the total number of instructions and $S$ denotes the G-eval score.
\textbf{Error + Coordination} (0.22) is highly effective, indicating that
integrated body coordination feedback ensures athletes adjust their full-body
mechanics rather than isolated actions. \textbf{Error + Method} achieves the
highest contribution (0.3), as identifying errors while providing corrective
strategies significantly assists refining movements. For instance, in
Fig.~\ref{fig:Demo}-A, 
the issue is effectively
pinpointed and clear corrective instruction is suggested. By contrast,
\textbf{Time + Body Part} (0.07) is less effective as no actionable guidance is
given. 

Figure~\ref{fig:combination_analyze} illustrates the accuracy of instructions
which meet two indicators. Dark color represents a high average G-eval score.
With more darker areas, results show that CoachMe generates more accurate
instructions. Moreover, CoachMe generates more coordination-related
instructions, which confirms that the proposed Human Pose Perception
(Sec.~\ref{sc:human pose perception}) enables CoachMe to analyze
full-body movement. Particularly, CoachMe achieves relatively high scores on
all dual indicators, showing that the instructions it generates consider multiple
aspects like experts. 
Overall, providing a precise solution in and of itself helps greatly; further 
identifying the problem while suggesting a solution only improves the
instruction. Findings~1 and~2 suggest the superiority of CoachMe from
these aspects.


\subsection{Finding 3: CoachMe Captures Sport-Specific Instructional Patterns}
\label{designOfCoachMe}
CoachMe demonstrates a strong ability to replicate the distribution of sport indicators found in ground-truth instructions annotated by professional coaches. Its predicted instruction distributions in both figure skating (FS) and boxing (BX) closely align with those of the actual datasets (Fig. \ref{fig:Skating_GT_bar}, \ref{fig:Boxing_GT_bar}, and \ref{fig:BX_FS_matrix}), indicating effective domain adaptation.
In the BX dataset, both instruction predicted by CoachMe and ground truth rarely mention sport indicator "Time Detection" and "Coordination." Expert feedback reveals that beginner-level boxing instruction favors simple, body-part-specific guidance over complex relational cues. In addition, coach also indicates that a basic punch video is very short, temporal errors detection is not useful in such motion. Fortunately, CoachMe learns this pattern well, aligning its outputs with the coaching style used for fundamental actions.
These findings underscore the critical role of domain-specific knowledge, given that each sport has its own distinct movement characteristics. This aligns with CoachMe’s core design principle—namely, its ability to flexibly adapt to various sports by learning sport-specific patterns through efficient, lightweight adaptation modules.

\section{Conclusion}
\par We propose CoachMe, a reference-based motion-to-instruction model that generates tailored instructions for sports. With the injection of concept difference, human pose perception, and instruct motion, CoachMe simulates a coach's thinking process, which identifies deviations from standard motions and focuses on both crucial body parts and their physical movements. Through comprehensive evaluations using quantitative performance metrics, human evaluation, and visual explanations on skating and boxing datasets, we demonstrate CoachMe's ability to provide high accuracy and good sport utility
in generated instructions for learners. 

\par CoachMe achieves state-of-the-art results for both motion description and instruction generation. Moreover, it represents a significant step forward to bridge the gap between artificial intelligence and human expertise in
athletic training. This light-weight, standalone, high-performance, adaptive model yields good opportunities for deployment in various sport scenarios.

%% file: latex/H_limitation/limitation.tex
\section*{Limitations}
\par CoachMe is primarily designed for beginner- and intermediate-level practitioners.
Our dataset consists of practice videos from novice athletes, and the generated instructions focus on fundamental movement correction. Therefore, CoachMe may not be well-suited to provide guidance on advanced techniques or professional-level performance. Future work could explore adapting the model to higher-level athletes by incorporating expert demonstrations and more complex movement evaluation.

\par In the practical implementation of CoachMe, we prioritize maintaining a consistent coaching style to authentically replicate the experience of working with a real coach. 
This approach ensures that users receive clear, coherent, and dependable guidance, as demonstrated throughout our paper.
During our research, interviews with athletes revealed a clear preference for personalized and consistent coaching styles, as opposed to general or mixed approaches.
Nevertheless, the real world offers diverse coaching styles, and each athlete has their own unique preferences.
To meet this demand, CoachMe is designed with a flexible architecture that allows the creation of new virtual coaches simply by gathering instructional datasets from real-world coaches.
This adaptability not only enables athletes to select virtual coaches aligned with their individual needs and preferences, but also empowers CoachMe’s scalability to seamlessly incorporate coaching data from a wide range of experts — including globally recognized professionals.
In summary, CoachMe currently learns only a specific teaching style based on our dataset, which limits its ability to represent the diverse speaking styles and instructional approaches of coaches in real-world scenarios. Future work could focus on incorporating a broader range of coaching styles to enhance its generalizability.
\par One limitation of CoachMe stems from the absence of figure skating and boxing data in the HumanML3D, which can lead to misclassification of actions. To address this, future work could explore integrating an additional action classification module to provide more explicit movement categorization. Furthermore, augmenting HumanML3D with sport-specific data—such as curated figure skating or boxing sequences—could significantly enhance CoachMe’s ability to accurately recognize subtle movement distinctions. By enriching the dataset with targeted sports movements and creating sport-specific versions of CoachMe tailored to the unique features of each discipline, we can continuously solve this issue.

\section*{Ethics Statement}
All participants whose videos are included in our dataset provided informed consent, acknowledging that their videos would be used for research purposes. For participants who are minors, we obtained consent forms from their parents or legal guardians. We ensured that all data collection and usage practices complied with ethical guidelines and respected the privacy and rights of the participants. Additionally, no personally identifiable information was revealed in the dataset, and all files were processed to ensure anonymity.

\section*{Acknowledgments}
This work is supported by the National Science and Technology Council of Taiwan under grants 114-2425-H-007-002 and 114-2425-H-007-001.
We extend our appreciation to Kristina Stepanova, a professional figure skating coach, for her insightful guidance on figure skating techniques, and for also providing expert human evaluation.
We also thank Shirley Tzeng for contributing figure skating movement data.
These movements and instructional descriptions were used to construct the Figure Skating (FS) dataset.
We also gratefully acknowledge the support of National Tsing Hua University, with special thanks to professional boxing coaches Kai-Chun Hong and Chia-Yao Chung for their expert annotations and contributions of instructional data on boxing movements. 
These movements and instructional descriptions were used to construct the Boxing (BX) dataset. We also thank Chen-Wei Pan for providing professional human evaluation.

%% file: latex/I_Appendix/appendix.tex
\appendix
\section{Dataset Details}
\subsection{Dataset Construction}
We divided the dataset into training and testing sets in a $4:1$ ratio. Given that a single figure skating video can be split into multiple video clips corresponding to error segment annotated by professional figure skating coach, we ensured that clips from the same video did not appear in both the training and testing sets. 
This approach was implemented to prevent data leakage and ensure that the model's performance evaluation is based on unseen data. Therefore, the number of training videos in the Figure Skating (FS) dataset is 292, with 64 videos for testing, as shown in Table \ref{sc:datasets}. The HumanML3D dataset has a frame rate of 20 frames per second (FPS), the Boxing (BX) dataset operates at 60 FPS, and the FS dataset is recorded at 30 FPS.
\subsection{Data Preprocessing}
To best mitigate noise originated from the background, for all rgb settings and videos we track all people in the scene using YOLOv7, manually select the main person index then resize the video into 224x224, centered at the main character.
\par In human pose perception, we employ HybrIK \cite{li2022hybrik} to predict 22 joint coordinates in a local coordinate system following the SMPL \cite{SMPL:2015} format for videos from the Figure Skating and Boxing datasets. To ensure consistency between the pretraining and finetuning settings, we localize the coordinates in the Human ML3D dataset. Localization involves subtracting the coordinates of the first index joint, which is the pelvis. After localization, the motion videos appear to rotate around the pelvis.

\subsection{Data Augmentation Template}
\label{Appendix:Data Augmentation Template}
In Section \ref{sc:datasets}, we incorporate the GPT-4 API to diversify the
original dataset. The template we used is shown in Table~\ref{tab:a} and Table~\ref{tab:b}.
\par Because the figure skating coach provides one or multiple error segments for each figure skating video, along with the corresponding original ground-truth instruction for each segment, we apply data augmentation accordingly.
Specifically, for each ground-truth instruction, we generate five augmented instructions using five different templates, as shown in Table \ref{tab:a}.
As a result, each annotated error segment, a ground-truth clip, has six associated instructions: one original and five augmented versions.
For the FS dataset, the augmented instructions maintain a high degree of similarity with their corresponding original instructions.
The average similarity score is $0.93$, with $100.00\%$ of the instructions scoring above $0.8$, and $92.40\%$ above $0.9$.
\par The augmented instructions for the BX dataset show even higher similarity scores with their originals. 
The average similarity score is $0.95$, with $100.00\%$ above $0.8$ and $98.38\%$ above $0.9$.
This is because each boxing video includes three ground-truth instructions annotated independently by three different coaches.
When applying augmentation to these three original instructions, we use the same template as shown in Table \ref{tab:b}.
As a result, each raw video (without coach-annotated error segmentation) yields six labels in total: three original and three augmented instructions.
\par Additionally, we appended a restrictive prompt—“... not begin with a directive verb...”—to the end of each prompt template. This decision stems from our observation that when models are asked to rephrase an instruction—which typically emphasizes keeping the tone positive, encouraging, and supportive—they often adopt an overly encouraging tone and frequently begin with directive verbs such as "Keep...".
To counter this tendency, we explicitly prohibit such language and instead require a neutral tone.
Furthermore, to minimize hallucination during data augmentation using GPT-4o, we also include the constraint: “Do not introduce any information that is not present in the target instruction.” This is designed to ensure that CoachMe learns from augmented data without incorporating hallucinated or fabricated content.
\begin{table}[ht]
\centering
\small
\begin{tabular}{ | l |p{6cm}  |}
\hline
Role & Content \\
\hline
system & You are an experienced figure skating coach who specializes in helping students improve their skating skills, particularly with the \{motion type\} jump.

Your task is to rephrase the instruction.

Please follow this guideline when rewriting:

Guideline: 
    \begin{enumerate}
        \item Use simple and clear language that beginners can easily understand and apply.  \
        \item Maintain a clear and neutral tone with a professional and objective style. \
        \item Feel free to omit parts of the original instruction that are not particularly helpful. \
        \item Avoid phrasing that sounds too strict or overly commanding. \
        \item Focus on offering constructive suggestions that help students feel motivated to improve. \
    \end{enumerate}
    
If the target instruction does not begin with a directive verb such as "Keep", "Try", "Focus on", "Ensure", "Consider", "Aim", "Avoid", "Make sure", or "Remember", you should avoid introducing one in the rephrased version. Maintain a neutral tone. Do not introduce any information that is not present in the target instruction. 

Please provide exactly one alternative way to rephrase the instruction. \\
\hline
user & Target instruction: \{instruction\} \\
\hline
\end{tabular}
\caption{Data augmentation template for GPT-4, where \{motion type\} refers to the
current jump type (Axel, Lutz, Loop) and \{instruction\} refers to
the original annotation.}
\label{tab:a}
\end{table}

\begin{table}[ht]
\centering
\small
\begin{tabular}{ | l |p{6cm}  |}
\hline
Role & Content \\
\hline
system & You are an experienced boxing coach who specializes in helping students improve their boxing skills, particularly with the \{motion type\} technique.

Your task is to rephrase the instruction.

Please follow this guideline when rewriting:
Use simple and clear language that beginners can easily understand and apply.

Guideline: 
    \begin{enumerate}
        \item Use simple and clear language that beginners can easily understand and apply.  \
    \end{enumerate}
    
If the target instruction does not begin with a directive verb such as "Keep", "Try", "Focus on", "Ensure", "Consider", "Aim", "Avoid", "Make sure", or "Remember", you should avoid introducing one in the rephrased version. Maintain a neutral tone. Do not introduce any information that is not present in the target instruction. 

Please provide exactly one alternative way to rephrase the instruction. \\
\hline
user & Target instruction: \{instruction\} \\
\hline
\end{tabular}
\caption{Data augmentation template for GPT-4, where \{motion type\} refers to the
current boxing technique (Jab and Cross) and \{instruction\} refers to
the original annotation.}
\label{tab:b}
\end{table}

\subsection{Human Annotation}
We employed human annotators for data labeling, with skating annotations provided by professional coaches from Europe and Asia, and boxing annotations conducted by members of a college boxing team. Annotators were informed that their labeled data would be used to train models. Annotators were recruited based on their domain expertise and compensated at a rate of \$50 per hour. The data collection protocol was reviewed and approved by the Institutional Review Board (IRB).

\input{latex/I_Appendix/PoseEstimator}

\section{Model}
\subsection{Hyperparameter Settings}
\par For the optimizer, we use AdamW with a learning rate of 1e-4 and 5000 warm-up steps for both the pre-training and fine-tuning settings.
\par For pretraining, we used an A100 GPU, with a total training time of 1720 minutes ($17$ min $12$ sec $\times 100$). The batch size was set to 16, with a maximum of 50 epochs.
\par For finetuning, we set the learning rate to $1 \times 10^{-4}$ with a maximum of 200 epochs and 5000 warmup steps. The batch size is set to 4. LoRA configurations for Human Pose Perception and Projection use both $\text{bias} = \text{none}$, $r = 32$, $\text{alpha} = 64$, and $\text{dropout} = 0.1$. The dropout rate is set to 0.5 for Human Pose Perception and 0.1 for the projection layer. Training was carried out on an RTX 4090, boxing training took 500 minutes ($2.5 \times 200$) and skating training 200 minutes ($1 \times 200$). 

\begin{table}
\small
\centering
\begin{tabular}{c c c c}
\hline 
Module & FS & BX \\ \hline
Motion Alignment & 0.37sec & 0.30sec \\ \hline
HyBrIK & 21sec  & 35sec \\ \hline
Human Pose Perception \\
\& InstructMotion & 1.93sec & 2.09sec \\
    \hline
    \end{tabular}
    \caption{The inference times of Motion Alignment, HyBrIK, and Human Pose Perception on the Figure Skating (FS) and Boxing (BX) datasets.}
    \label{table:inference time}
\end{table}
As shown in \ref{table:inference time}, we can see HybrIK preprocessing accounts for 90\% of the inference time, while CoachMe only requires the remaining 10\%.

\begin{table}
\small
\centering
\begin{tabular}{c c c c}
\hline 
Module & \# of Parameters   \\ \hline
Basic CoachMe & 305.54M \\ \hline
Lora of Figure Skating & 5.64M \\ \hline
Lora of Boxing & 5.64M  \\ \hline
Trainable parameters\\ of CoachMe & 232.88M \\ \hline
T5-base & 223M \\ 
    \hline
    \end{tabular}
    \caption{The number of parameters in Basic CoachMe and the trainable parameters in CoachMe for sport-specific tasks.}
    \label{table:inference time}
\end{table}
\par The number of parameters in Basic CoachMe and the trainable parameters in CoachMe for sport-specific tasks.

\par For Human Pose Perception, we set the number of kernels in the convolutional neural network to 1024. This is to capture diverse motion details in motion descriptions and various error types in motion instructions.

\subsection{Model Configuration}
This section presents a detailed layer-wise analysis of each model's architecture, elaborating on the structure and functionality of individual components. Table 5 provides a comprehensive overview of the parameter counts along with the corresponding architectural configurations for all modules.
\begin{table*}[bt]
\small
\centering
\begin{tabular}{c c c c c c}
\hline 
\multicolumn{6}{c}{\textbf{Concept Difference}} \\ \hline
\multicolumn{6}{c}{Concept Encoder } \\ \hline
Module & ResNet-50 (finetune) & MLP & Transformer Encoder (x3) & Final Projection \\
\# param. & 23.5M & 1.7M & 3.1M & 0.13M \\ 
shape & →2048 & →256 & →256 & →128 \\ \hline
\multicolumn{6}{c}{Error Segment Identification} \\ \hline
Module & Input FC & Self-Attention & FFN &	Layers Norm & Output FC \\
\# param. & 2K & 1K & 65K & 64 & 34 \\
shape &	16->16 & ->2048 & ->16 & ->16 & ->2 \\ \hline 
\multicolumn{6}{c}{\textbf{Human Pose Perception}} \\ \hline 
\multicolumn{6}{c}{Pose Understanding} \\ 
Block & 0 & 1 & 2 & 3 & 4   \\ 
\# param. & 14K & 20K & 58K & 70K & 22K \\ 
shape & 6->32 & ->32 & ->32 & ->64 & ->128\\ \hline
\multicolumn{6}{c}{Pose Extraction} \\
Block & 0 & 1 & 2 & 3 & 4   \\ 
\# param. & 267K & 267K & 859K & 1M & 1M \\ 
shape & 128->128 & ->128 & ->256 & ->256 & ->256 \\
Layer & BN & Conv & AttnJ & AttnG   \\ 
\# param. & 512 & 262K & 1K & 90K \\ \hline
\multicolumn{6}{c}{Pose Attention} \\ \hline
Block & 0 & 1 & 2 & 3 & 4   \\ 
\# param. & 172K & 172K & 645K & 636K & 636K \\ 
shape & 128->128 & ->128 & ->256 & ->256 & ->256 \\ \hline
\multicolumn{6}{c}{\textbf{Instruct Motion}}\\ \hline
\multicolumn{6}{c}{Projection} \\ \hline
Layer & 0 & 1 & 2 & 3   \\
\# param. & 51K & 34K & 34K & 42K \\
shape & 512->512 & ->512 & ->512 & ->768 \\ \hline
\multicolumn{6}{c}{Language Model : T5-base}\\ \hline
\# param. & 223M \\
    \hline
    \end{tabular}
    \caption{Layer-wise architectural configurations and parameter counts of all models. \# param. is denoted as the number of parameters. BN is denoted as the batch normalize. Conv is denoted as the convolution. AttnJ is denoted as the attention joint. AttnG is denoted as the attention graph.}
    \label{table:inference time}
\end{table*}

\subsubsection{Concept Difference}
\par The Concept Encoder consists of four main components: a ResNet-50 backbone with partial finetuning, a lightweight MLP transformation module, a 3-layer Transformer Encoder, and a final projection head. The ResNet-50 backbone is finetuned from stage 4 onward, transforming the input RGB clip into a 2048-dimensional feature representation. This is followed by an MLP comprising three fully connected layers that reduce the dimensionality to 256. The Transformer Encoder further processes the temporal sequence of features across three layers, each with multi-head self-attention and feed-forward blocks. The final projection head maps the representation to a 128-dimensional embedding space for motion alignment.

\par For the Error Segment Identification module, we adopt a simple transformer-style architecture. The input first passes through a linear layer to expand the feature space from 16 to 16 dimensions, followed by a single-layer self-attention mechanism. This is followed by a feed-forward network with an intermediate dimensionality of 2048. Layer normalization is applied before and after the attention and feed-forward layers. A final output layer maps the sequence to a binary prediction indicating whether a frame belongs to an error segment.
\subsubsection{Human Pose Perception}
\par In Human Pose Perception, Pose Understanding, Pose Extraction, and Pose Attention all consist of five identical blocks. Each block includes a spatial convolution, implemented as a single 2D convolution layer, followed by a temporal convolution composed of a batch normalization layer, a ReLU activation function, a 2D convolution layer, another batch normalization layer, and a second ReLU activation function, applied sequentially. Additionally, each block incorporates a residual connection that consists of a 2D convolution layer followed by a batch normalization layer.
\par After the Pose Extraction blocks, the output is first processed by a batch normalization layer followed by a convolution layer to produce the attention embedding. The attention embedding is then further processed sequentially by a convolution layer, a batch normalization layer, and a linear interpolation operation. Finally, a sigmoid activation function is applied to generate the attention joint. In parallel, the attention embedding first undergoes average pooling, convolution, and batch normalization, then passes through tanh and ReLU activations to generate the attention graph.
\par In terms of the graph, except for Pose Attention, which utilizes the attention graph, the rest are based on the spatial graph.
\subsubsection{Instruct Motion}
\par The Projection applies temporal average pooling, followed by three consecutive blocks, each consisting of a linear layer, a batch normalization layer, and a ReLU activation function applied in sequence. Then, a final linear layer to map from 512 to 768 dimensions for T5-base input. 
\par We choose beam search as our decoding strategies in InstructMotion, as discussed in Section \ref{sc:Instruction Generation}. As beam search typically generates better descriptions of motion sequences by considering more candidate options, we have chosen it over greedy search. We set the beam size of beam search is 3.
\par For tokenizer, we utilize the AutoTokenizer from the Hugging Face Transformers library to efficiently process input sequences. This ensures optimized tokenization while maintaining compatibility with the T5-base model. For the T5 configuration, we use T5ForConditionalGeneration from the Hugging Face Transformers library to initialize the model with a predefined configuration. This guarantees consistency in the model architecture and parameter settings, while leveraging the pretrained weights of T5-base for effective sequence-to-sequence generation.

\subsection{Pretrain and Finetune}
We adopt the idea of Chain of Thought \citep{lee2023teaching}. We use "Motion Description" during pretrain and "Motion Instruction" during finetune as the start tokens separately. Therefore, model can distinguish whether to generate motion description or instructional language according to the task prompt given from the start tokens. 

\section{Visualization}
\subsection{t-SNE Visualization of Learner-Standard Pair}
\begin{figure*}[ht]
    \centering
    \includegraphics[width=\textwidth]{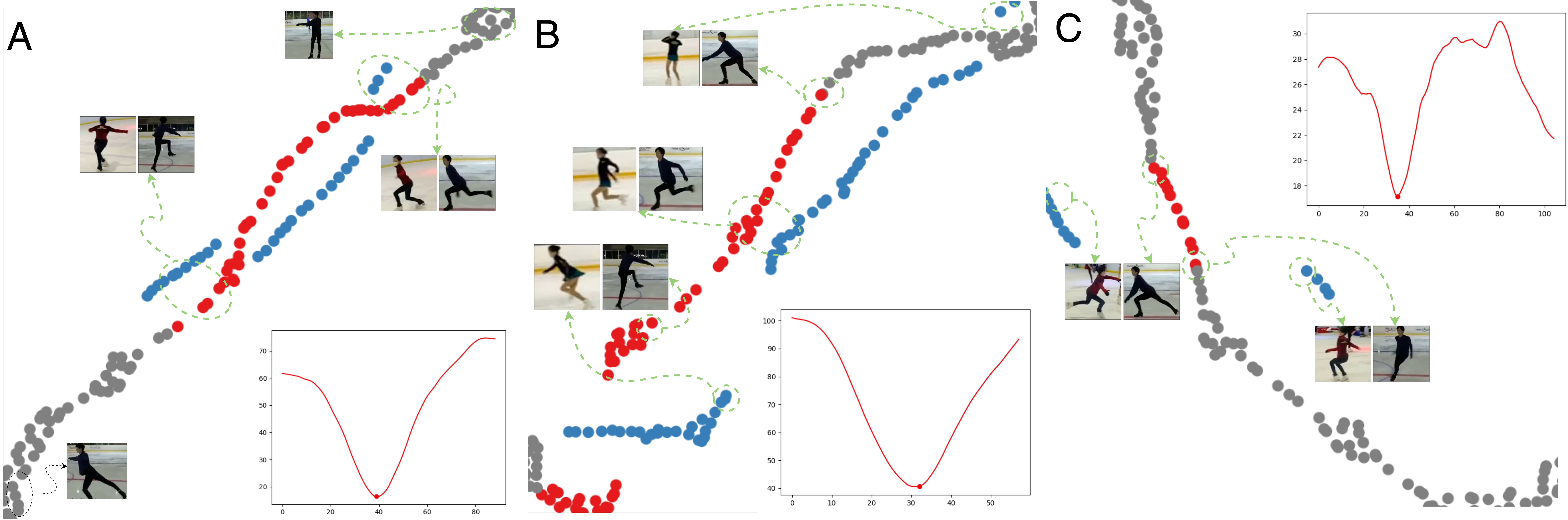}
	 \caption{When possible, we mark the same frame using a single circle and
	 link it to the frame comparison. If this is not possible, two circles are
	 drawn to indicate their corresponding frames.}
    \label{fig:appendix}
\end{figure*}

In Figure ~\ref{fig:appendix} we present more examples of selecting the minimum distance and visualized 
embeddings.
\par In illustration \textbf{A}, we show that the gray frames are not selected since
the frame in the standard video represents preparation for the Axel, which is
not included in the input video. Illustrations~\textbf{B} and~\textbf{C}
showcase the aligned start and end frames by computing the minimum distance
using the DTW cost matrix. Each illustration is associated with its
corresponding DTW cost matrix, where the red dot denotes the minimum value.
\subsection{Visualize Attention of Human Pose Perception}
\label{Human Pose Perception}
To explain why certain motion elements are generated in the instructions, we visualize the attention graph and joints used in Human Pose Perception. As shown in Figure \ref{fig:Demo}, We visualize the top-3 important relationships of two joints, which is recorded in attention graph, and top-3 important joints, which is recorded in attention joints, that learned by Human Extraction. We set the number of attention graph to $4$. Figure \ref{fig:Demo} shows that the different attention graphs can pay attention on different human body-parts.

\section{Evaluation}
\subsection{Metrics}
\label{sec:Metrics}
Following \citet{guo2022tm2t} and \citet{jiang2024motiongpt}, we utilized several
standard metrics: BLEU~\cite{papineni-etal-2002-bleu}, 
ROUGE~\cite{lin-2004-rouge}, and BertScore~\cite{zhang2020bertscore}. 
\par However, BLEU scores generally evaluate the similarity and identification of word spans. By nature, these scores tend to be higher in some generation tasks where the generated texts are more consistent, and lower in others. In this paper, we observe the same trend: the BLEU scores for motion descriptions \ref{tab:motion_description_comp} are higher than those for generated instructions \ref{tab:combined_results}. Model reliability is not the reason for this difference. The human evaluation in \ref{table:win_rate} further supports the performance superiority of the proposed model.
\par Therefore, to better assess the quality of generated text instructions, we incorporated 
G-Eval~\cite{liu2023gevalnlgevaluationusing} to evaluate the consistency between the
predicted instructions and the ground truth. The scale was between $1$ to $5$. This
was done by prompting Claude to evaluate the score of consistency with a dedicated template as follows (See. Sec \ref{Appendix:G-Eval Template}).
We followed the original paper and prompted Claude five times to avoid ties in scores. Additionally, we consulted domain experts to identify subtle differences that can
be discerned only by specialists in the field.

\subsection{Detailed Overview of Six Sport Indicators}
\label{Appendix:Indicator_Detail}
In Section \ref{Disscussion:Instruction Indicators}, we define six key indicators, each of which is described in detail below.
The evaluation of whether the instruction contain these factors is based on the design of prompts in 
reference to the G-eval template~\cite{liu2023gevalnlgevaluationusing}, as detailed in Section \ref{Appendix:G-Eval Template}.

\textbf{(1)~Detecting errors in the motion}~-- Identifying mistakes such as
improper posture, imbalance, or misalignment.

\textbf{(2)~Identifying timing information}~-- Determining if an error occurs
\textbf{during takeoff, mid-air, or landing}, or if there is a \textbf{rhythm
disruption}.

\textbf{(3)~Recognizing body part movements}~-- Pointing out the specific body
parts involved in an error, such as the left knee dropping or the right
shoulder tilting.  

\textbf{(4)~Identifying causal relationships in the sport}~-- Explaining how an
error affects performance, such as instability from poor foot positioning.

\textbf{(5)~Explaining how to improve the sport}~-- Providing corrections such
as lifting the right leg higher for better balance.

\textbf{(6)~Describing how body parts coordinate or interact}~-- Explaining how
multiple body parts should work together or interact, such as aligning the hips
and shoulders during rotation.

\subsection{Sport Indicators on FS and BX datasets}
\begin{figure}[t]
  \includegraphics[width=1\columnwidth]{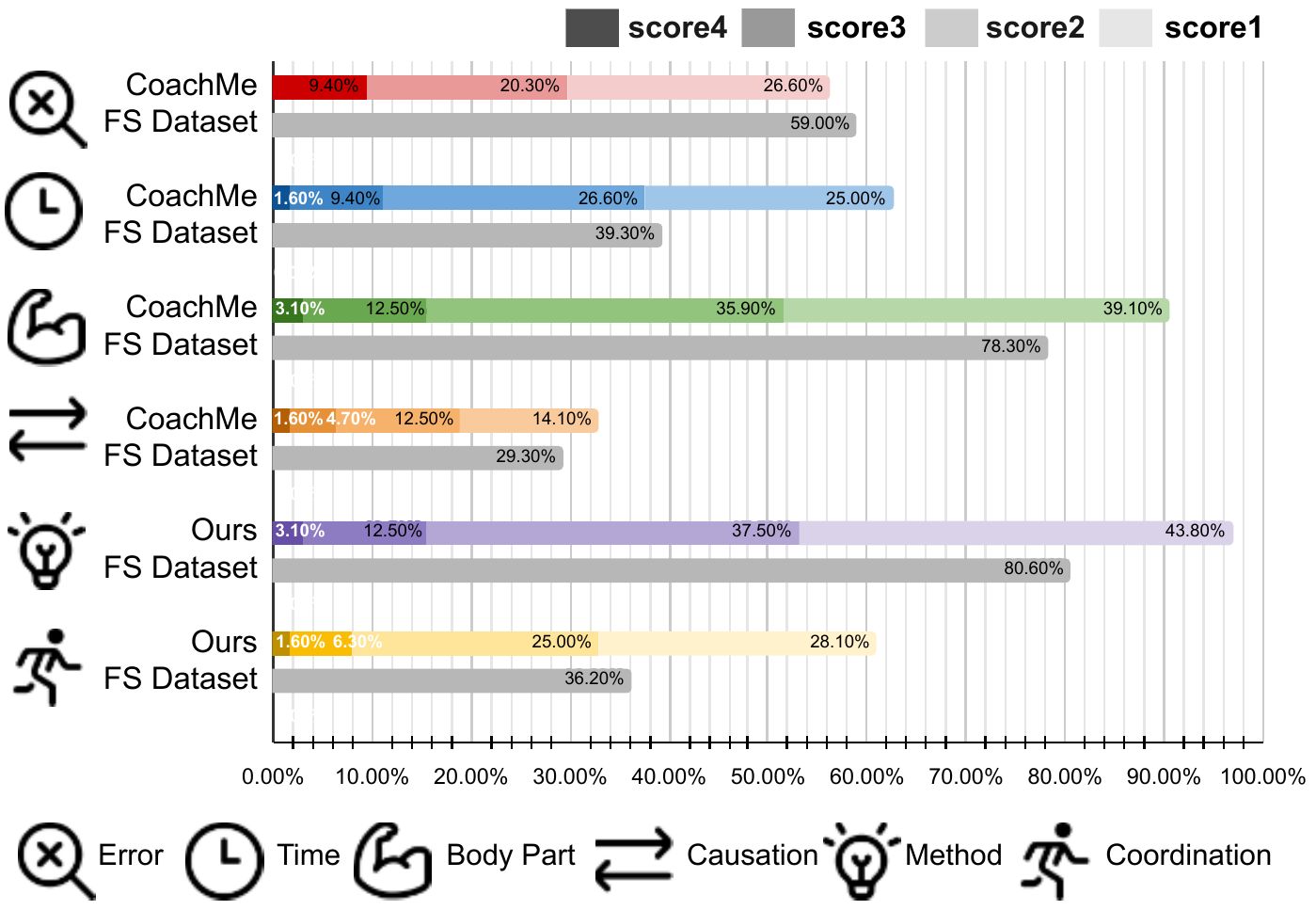}
  \caption {The horizontal bar chart represents the to-
tal number of predicted figure skating instructions of CoachMe assigned to each indicator (Sec. 4.4), with colors indicating their respective G-eval scores. The gray bar is the ground truth boxing instruction in FS dataset.}
   \label{fig:Skating_GT_bar}
 \end{figure}
 \begin{figure}[t]
  \includegraphics[width=1\columnwidth]{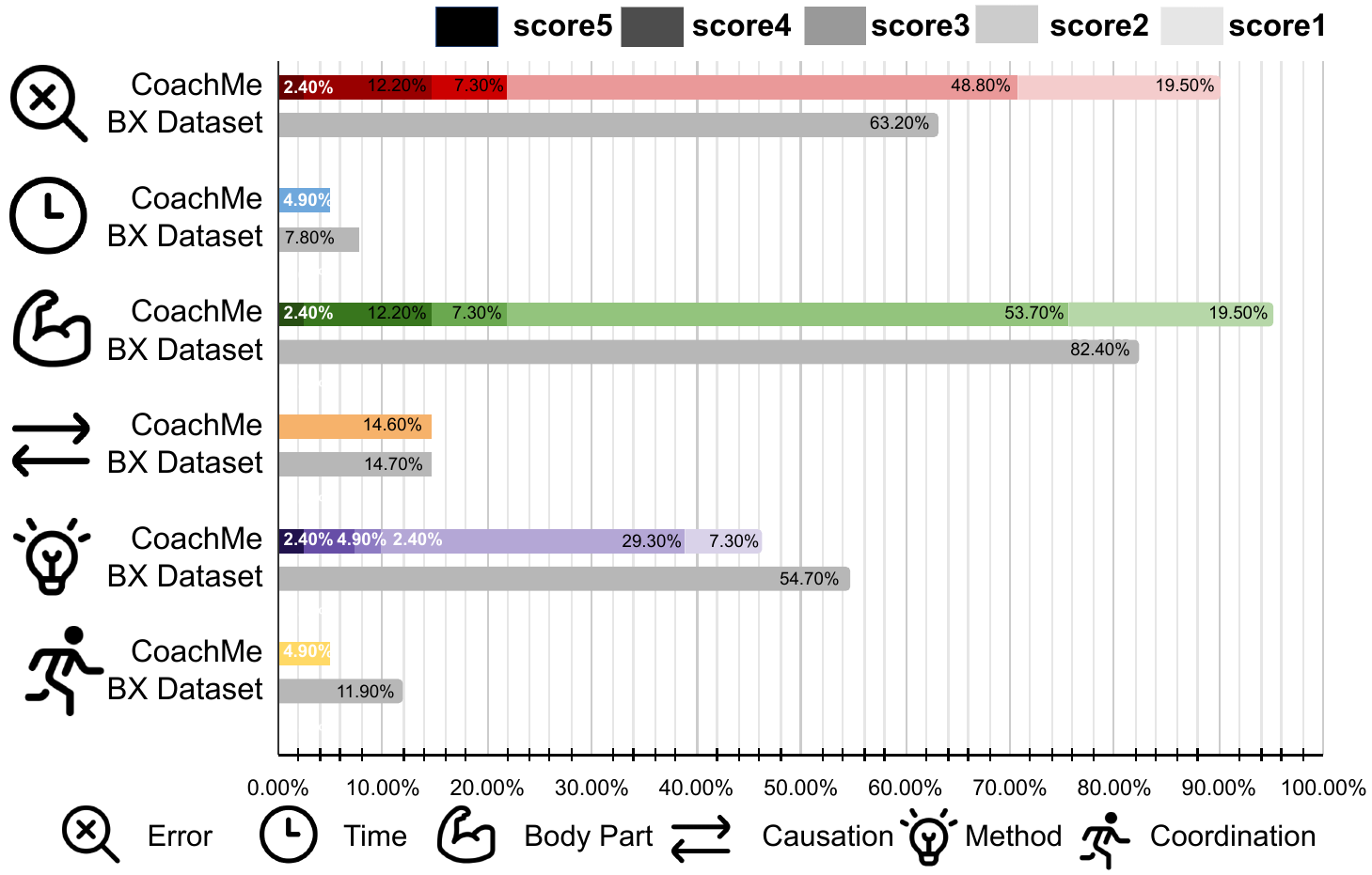}
  \caption {The horizontal bar chart represents the to-
tal number of predicted boxing instructions of CoachMe assigned to each indicator (Sec. 4.4), with colors indicating their respective G-eval scores. The gray bar is the ground truth boxing instruction in BX dataset.}
   \label{fig:Boxing_GT_bar}
 \end{figure}
In addition to analyzing the instructions predicted by CoachMe, LLaMA, and GPT-4o, we also examined the distribution of sport indicators in the ground-truth instructions annotated by professional coaches. The differences in sport indicator distributions between figure skating and boxing, as shown in the Figure \ref{fig:Skating_GT_bar} and Figure \ref{fig:Boxing_GT_bar}, highlight the significant distinctions between these two sports. In both figures, we can also observe that the distribution of instructions predicted by CoachMe is almost consistent with those of FS and BX.
\par CoachMe achieves G-eval consistency scores of 2.20 and 1.83 with respect to the ground-truth instructions in the figure skating (FS) and boxing (BX) datasets, respectively. These scores indicate that, while there remains a noticeable gap in semantic similarity to the ground-truth instructions, CoachMe successfully captures and aligns with the sport-specific instructional patterns. This is evident in its sport indicator distributions, which closely mirror the domain-specific patterns in both sports. These results suggest that CoachMe is capable of generating sport-specific instruction.

\begin{figure}[t]
  \includegraphics[width=1\columnwidth]{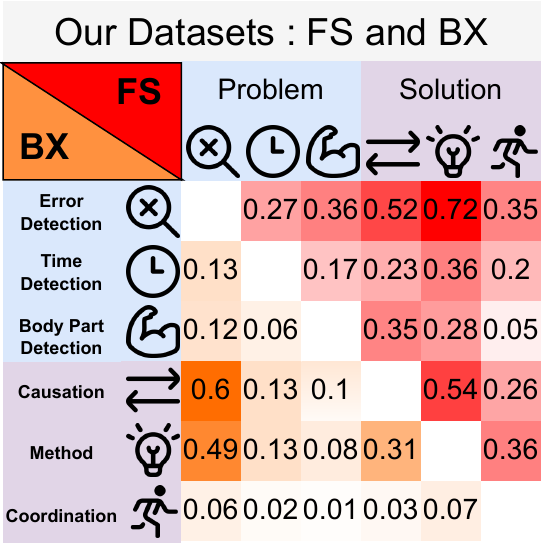}
  \caption {Effectiveness of indicator combinations on the FS and BX datasets.
  In the evaluation matrix, red numbers in the upper-right triangle represent scores on the FS dataset.
  Orange numbers in the lower-left triangle show scores on the BX dataset.
  Each number denotes the percentage $P$ of the ground truth that contain two indicators, calculated based on Eq. \ref{eq:percenrageGT}}
   \label{fig:BX_FS_matrix}
 \end{figure}
We also use six sport indicators to analyze our ground truth instruction in the FS and BX datasets, as shown in Figure \ref{fig:BX_FS_matrix}. The numbers in Figure \ref{fig:BX_FS_matrix} indicate the frequency with which the guidance prompts contain the indicators corresponding to both the row and the column. 
\begin{equation}
    \label{eq:percenrageGT}
    P = \frac{\mbox{\small number of instructions containing two indicators}}{N},
\end{equation}
where $N$ denotes the total number of instructions. 
After comparison with Figure \ref{fig:combination_analyze}, we can find that the distribution of the FS and BX datasets shows a striking similarity to the distribution of instructions predicted by CoachMe for figure skating and boxing videos.

\begin{figure}[bt]
    \centering
	 \includegraphics[width=0.6\columnwidth]{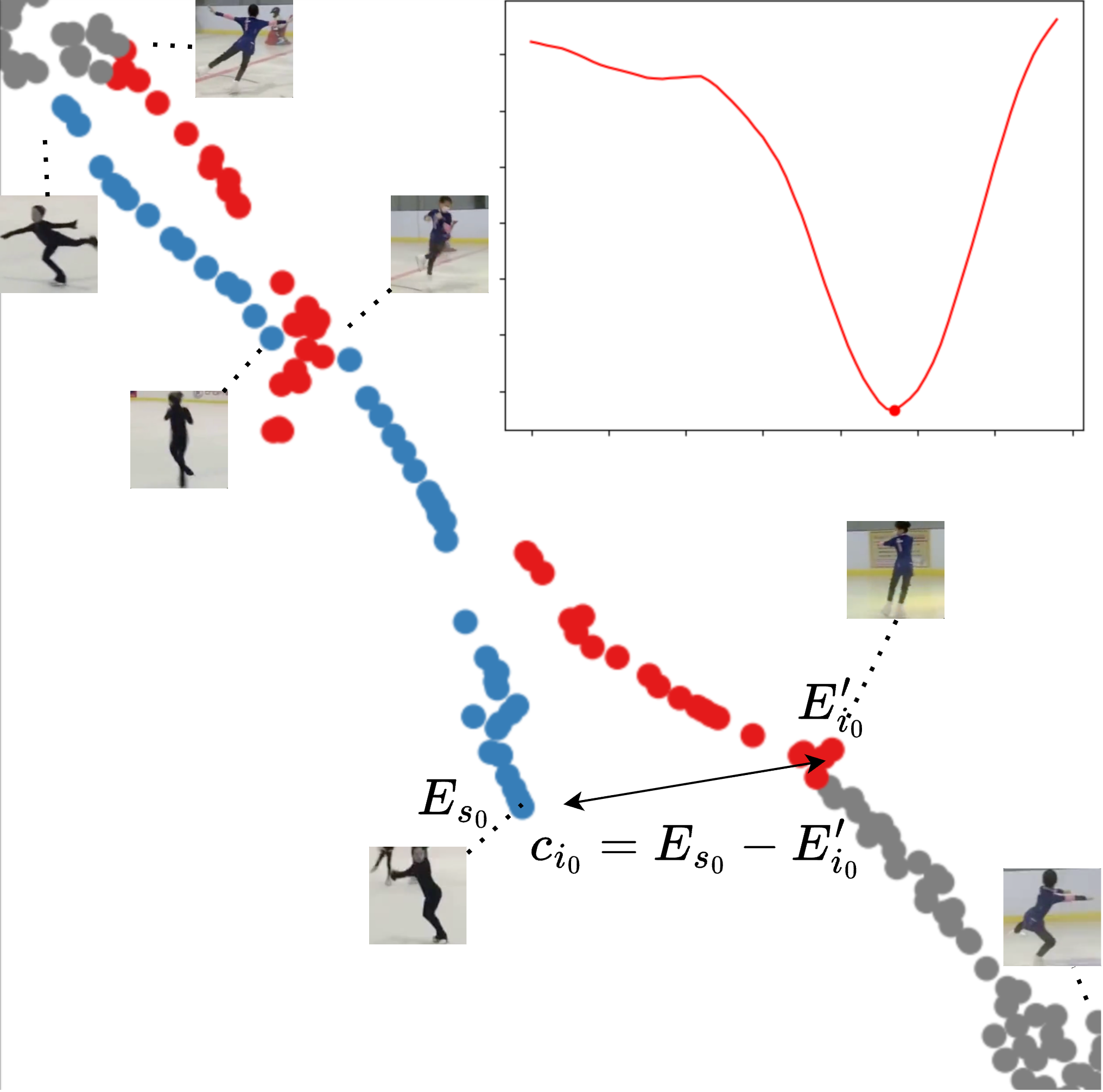}
	 \caption{Target motion retrieval (red) from input video using reference
	 video (blue). Top right: alignment cost for each starting frame. Gray dots
	 indicate frames outside the selected segment.}
    \label{fig:ss_example_squeeshed}
\end{figure}

\subsection{Comparison of Modalities in Human Evaluation}
\label{Appendix:Modalities_Human_Evaluation}
\begin{table}
\small
    \centering
    \begin{tabular}{c c c}
    \hline   
    Reference type & Best (\%) & Worst (\%)  \\
    \hline
    Basic CoachMe & 28.2 & \textbf{25}\\
    CoachMe & \textbf{48.4} & 36 \\ 
    CoachMe (RGB) & {23.4} & 39 \\ 
    \hline
    \end{tabular}
    \caption{Win rate of each CoachMe setting in human evaluation.}
    \label{table:Modalities_Human_Evaluation}
\end{table}
In this paper, we also collaborate with figure skating coaches to provide professional evaluations comparing the three configurations of our model: Basic CoachMe, CoachMe (RGB), and CoachMe (Baseline). Table~\ref{table:Modalities_Human_Evaluation} illustrates the results. Basic CoachMe
and CoachMe (RGB) settings fail to provide key insights for improving jumps. They
frequently generate guidance that always applies, such as ``Try to keep
your body straight while in the air,'' while overlooking other more critical
issues. Instead, CoachMe provides precise and personalized instructions. 
Notably, CoachMe generates instructions most frequently rated as the best. These results
align with the quantitative evaluation results shown in Table~\ref{table:win_rate}. It also validates that using skeleton as the motion token modality achieves superior performance in skating.

\subsection{Sport Indicators and Negative Factors in Human Evaluation}
\begin{figure}
  \includegraphics[width=1\columnwidth]{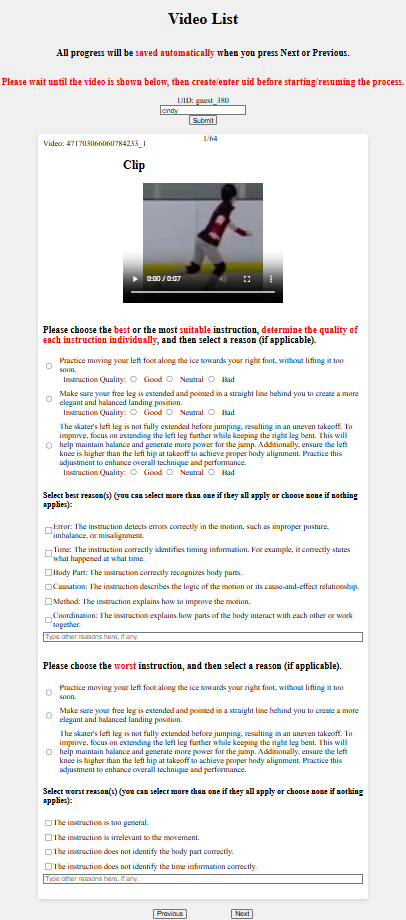}
  \caption {A sample video and its generated responses from three different
  models shown on the \ref{app:AOHE}.}
  \label{fig:web}
\end{figure}
To further analyze the effectiveness of generated instructions, during human evaluation by coaches, when a coach rated a model-generated instruction as the best or worst, they were also asked to specify the reasons for their selection, as shown in Figure \ref{fig:web}. The six reasons (See Sec. \ref{Disscussion:Instruction Indicators} and Sec. \ref{Appendix:Indicator_Detail}) why an instruction is rated as best correspond exactly to the six sport indicators. Therefore, we can examine the key indicators that contributed to high-quality feedback in human evaluation.
The four reasons why an instruction is rated as worst are the following below:

\textbf{(1) General}~-- The instruction is too general.

\textbf{(2) Irrelevant}~-- The instruction is irrelevant to the movement.

\textbf{(3) Incorrect Body Part}~-- The instruction does not identify the body part correctly.

\textbf{(4) Incorrect Time}~-- The instruction does not identify the time information correctly.

Professional coaches are allowed to select more than one reason for rating an instruction as best or worst if multiple reasons apply, or to choose none if none are appropriate.

\label{Appendix:Indicaotrs_Human_Evaluation}
\begin{table}
\small
    \centering
    \begin{tabular}{c c c c}
    \hline 
     reasons & GPT-4o  & LLaMA & CoachMe \\ \hline
    \multicolumn{4}{c}{\textbf{Figure Skating (FS)}} \\ \hline
     Rated as Best (\%) & 31.3 & 35.9 & 32.8 \\ \hline
     Error (\%)& 12.5 & 10.9 & 10.9 \\
     Time (\%) & 12.5 & 14.1 & 20.3 \\
     Body Part (\%) &  9.4 & 14.1 & 18.8 \\
     Causation (\%) & 17.2 & 15.6 & 18.8 \\
     Method (\%) & 15.6 & 10.9 & 21.9 \\
     Coordination (\%) & 3.1 & 4.7 & 12.5 \\ \hline
     Rated as Worst (\%) & 34.4 & 31.3 & 34.4 \\ \hline
     General (\%) & 23.4 & 35.0 & 25.0 \\
     Irrelevant (\%) & 10.9 & 18.8 & 12.5 \\
     Incorrect Body Part (\%) & 9.4 & 14.1 & 17.2  \\
     Incorrect Time (\%) & 4.7 & 6.3 & 3.1 \\ \hline
    \multicolumn{4}{c}{\textbf{Boxing (BX)}} \\ \hline
     Rated as Best (\%) & 26.8 & 39.0 & 34.1 \\ \hline
     Error (\%) & 42.1 & 71.4 & 56.5 \\
     Time (\%) & 31.6 & 57.1 & 43.5 \\
     Body Part (\%) & 21.1 & 47.6 & 39.1 \\
     Causation (\%) & 36.8 & 38.1 & 26.1 \\
     Method (\%) & 21.1 & 38.1 & 30.4 \\
     Coordination (\%) & 31.6 & 57.1 & 21.7 \\ \hline
     Rated as Worst (\%) & 39.0 & 39.0 & 22.0 \\ \hline
     General (\%) & 60.0 & 40.0 & 44.4 \\
     Irrelevant (\%) & 33.3 & 20.0 & 22.2 \\
     Incorrect Body Part (\%) & 53.3 & 20.0 & 11.1 \\
     Incorrect Time (\%) & 0 & 0 & 0 \\
    \hline
    \end{tabular}
    \caption{
    Comparison of the proportions of best/worst reasons for each model on Boxing (BX) dataset, based on 41 test videos and their corresponding predicted instructions. The percentages indicate how often an instruction was rated as best or worst due to the presence of the corresponding sport indicators or reasons.
    Error denotes error identification. Time denotes timing recognition. Body Part denotes body part awareness. Causation means causal relationships. Method means corrective methods. Coordination means coordination analysis. (See Sec. \ref{Disscussion:Instruction Indicators} and Sec. \ref{Appendix:Indicator_Detail}).}
    \label{table:reasons}
\end{table}
\begin{figure}[t]
\includegraphics[width=1\columnwidth]{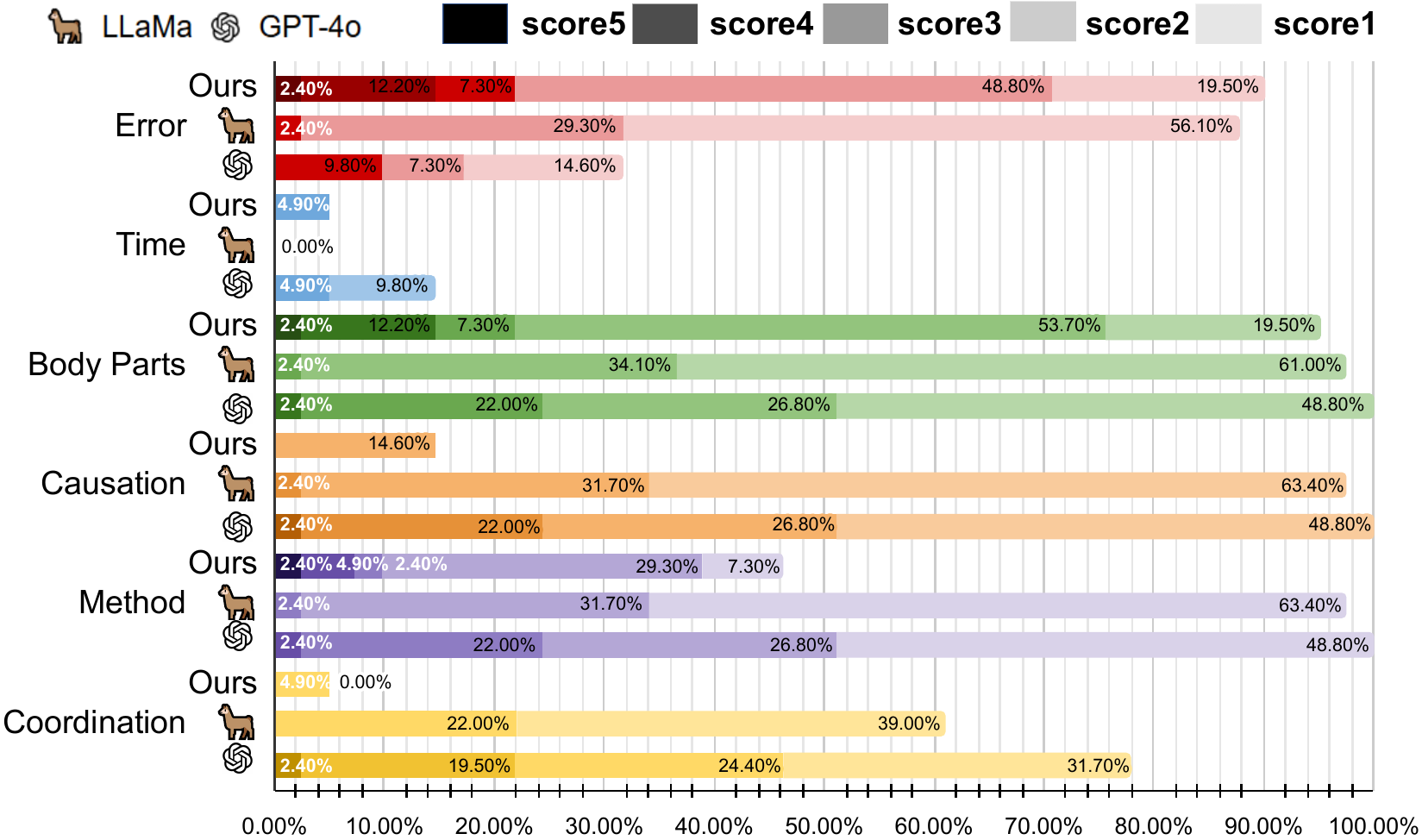}
  \caption {The horizontal bar chart represents the total number of
  instructions assigned to each indicator (Sec.~\ref{Disscussion:Instruction
  Indicators}), with colors indicating their respective G-eval scores. Higher
  G-eval scores indicate more informative and effective guidance that aligns
  more closely with the ground truth. The G-eval scores range from 1 to 5.}
   \label{fig:boxing_bar}
 \end{figure}
\subsection{Analysis of Human Evaluation on Figure Skating}
\label{FS-human_evaluation-details}
\par We analyzed the proportion of each sport indicators when a model was rated good, revealing how different models emphasize key instructional elements. Table~\ref{table:reasons} presents the distribution of $6$ sport indicators across GPT-4o, LLaMA, and CoachMe on Figure Skating (FS) dataset.
Among the six indicators, CoachMe consistently demonstrated superior performance, particularly in timing recognition (20.3\%), body part awareness (18.8\%), causal relationships (18.8\%), and corrective methods (21.9\%). This suggests that CoachMe not only identifies movement errors but also provides more actionable and context-aware feedback. Notably, coordination analysis, which is crucial for complex sports movements, was significantly higher in CoachMe (12.5\%) compared to GPT-4o (3.1\%) and LLaMA (4.7\%).

\par Regarding the quality of the generated instructions, we consulted professional figure skating coaches to obtain expert assessments. Overall, the coaches expressed strong appreciation and were notably impressed by the quality of instructions produced by CoachMe. Despite this positive reception, we conducted a further analysis of the subset of instructions ($29.7$\%) that received "bad" ratings in the human evaluation, as shown in Table \ref{table:win_rate}.

After discussions with coaches, we identified three common issues that contributed to the rating of an instruction as low quality on CoachMe.

\textbf{(1) Misidentification of jump types ($30.4$\%):} CoachMe identifies errors related to a specific jump type-for example, describing a toe loop when the actual jump performed is an Axel.

\textbf{(2) Correct but generic instruction ($21.7$\%):} The predicted instruction is technically accurate but fails to highlight the most critical technical detail that requires correction. 

\textbf{(3) False positive instructions ($30.4$\%):} The instruction correctly identifies a specific error, but the athlete’s actual execution is technically sound, leading to unnecessary corrections.

\par The first issue arises from our pretrained model trained on the HumanML3D dataset, which lacks detailed examples of specialized figure skating jumps. While these jumps are technically distinct, their biomechanical similarities make them difficult to differentiate. This limitation is more pronounced in general models like GPT-4o and LLaMA, which are not designed to capture domain-specific biomechanical cues. 
\par The occurrence of the second issue highlights the need for incorporating sport-specific knowledge into the instruction generation process. This problem is more prevalent in GPT-4o and LLaMA that often provide broadly accurate yet overly generic instructions, missing critical nuances (see Section \ref{Disscussion:accuracy} for detailed analysis). 
While the design of CoachMe aims to alleviate these issues, certain domain-specific inconsistencies persist.
This highlights the need for further refinement of our specialized models to more accurately identify and convey the most relevant movement details.
\par The third issue stems from a bias in our dataset, where all ground-truth instructions focus on suggesting improvements to the performed actions. This distribution leads CoachMe to implicitly assume that every input requires correction, even when the performance is already acceptable. To remedy this, we plan to introduce a scoring mechanism and integrate examples of high-quality movements with positive annotations. While related to the general challenge of hallucinations in large language models, we believe this issue can be effectively mitigated through future enhancements to our training data and model design.
\subsection{Analysis of results on Boxing}
\label{BX-human_evaluation-details}
The boxing coach who conducted the human evaluation also noted that since each clip features only a basic punch and is very short, temporal errors are not relevant. This explains why the "Incorrect Time" factor in Table \ref{table:reasons} consistently accounts for 0\% of the negative ratings.
As shown in Figure \ref{fig:boxing_bar}, We also observe that, across all models, instructions perform unusually poorly on the sport indicator "Time", suggesting that when the video is short—such as containing only a single punch—none of the models are able to provide meaningful temporal descriptions. This issue persists even in CoachMe, which specifically includes a Concept Difference Module designed to handle temporal information.

Despite this limitation, CoachMe outperforms both LLaMA and GPT significantly in the boxing domain across NLP metrics, G-eval, and human evaluations.

However, as shown in Figure~\ref{fig:boxing_bar}, CoachMe performs relatively poorly on the sport indicator Coordination. To understand this, we consulted boxing experts, who pointed out that in current boxing videos, it is often sufficient to describe individual body parts rather than the relationships between them. As a result, instructions generated by CoachMe can still receive high scores even with a low mention rate of Coordination.

This is because for beginner-level learners in the BX dataset, it is more effective to provide instructions targeting single body parts rather than describing complex interrelations among different body parts. Overloading beginners with corrections involving multiple body parts and their coordination may increase the complexity of the movements, making them harder to execute correctly. Therefore, professional coaches tend to focus on adjusting one specific body part at a time. This is particularly relevant in fundamental actions such as "Jab" and "Cross" or in stance training, which are designed to help novice boxers understand how to protect themselves during actual sparring.

This observation is reflected in the low frequency of the sport indicator "Coordination" in both the ground-truth distribution of the BX dataset and the instruction predictions generated by CoachMe, as shown in \ref{fig:Boxing_GT_bar} and \ref{fig:BX_FS_matrix}. Despite incorporating a Human Pose Perception designed to model coordination of different body parts, CoachMe effectively learns this pattern—providing simple, targeted instruction that mirrors such coach style employed by professional coaches when training beginners.

This highlights the importance of domain-specific knowledge, as each sport involves unique movement patterns.
This is precisely the design philosophy behind CoachMe: the ability to adapt to different sports by effectively acquiring domain-specific knowledge, supported by lightweight task-specific adaptation models.
However, this also suggests that the instructions should not only be sport-specific, but the evaluation indicators, the sport indicators, may also need to be tailored for each sport. In the future, it may be beneficial to design sport-specific sets of indicators tailored to the characteristics of each sport and action type.

\subsection{Human Evaluation Interface}
Figure \ref{fig:web} demonstrates the human evaluation interface
used in the experiment. Our annotator is required to select the best and worst
answers generated by the models, providing an explanation for each choice. Note
that throughout the evaluation, the annotator is not provided any
additional information (e.g., ground truth). This ensures that the information
seen by the annotator aligns with the information available to our model.

\subsection{G-Eval Template}
\label{Appendix:G-Eval Template}
\input{G-eval/consistency}
\input{G-eval/error_detection_prompt}
\input{G-eval/time_detection_prompt}
\input{G-eval/bodypart_prompt}
\input{G-eval/causation_detection_prompt}
\input{G-eval/method_detection_prompt}
\input{G-eval/coordination_detection_prompt}
We follow the original G-Eval paper \cite{liu2023gevalnlgevaluationusing} to design tailored prompts for evaluating the generated instructions. Claude (via its API) is employed as the evaluation LLM in our implementation of the G-Eval methodology. We adopt the G-Eval "Consistency" template to assess how well the generated instructions align with the ground truth, as illustrated in Table~\ref{G-eval:Consistency}. We also design specific prompts to evaluate the quality of instructions based on the indicators we have defined, as described in Section \ref{Disscussion:Instruction Indicators}. We adopt 6 G-Eval templates to evaluate specific aspects of instructional content: error detection (See Table \ref{G-eval:Error}), time-related expressions (See Table \ref{G-eval:Time}), descriptions that mention specific body parts (See Table \ref{G-eval:BodyPart}), causation or logical explanations of motion (See Table \ref{G-eval:Causation}), methods for improvement (See Table \ref{G-eval:Method}), and coordination between body parts (See Table \ref{G-eval:Coordination}).

%% file: latex/I_Appendix/PoseEstimator.tex
\section{Pose Estimator}
\label{PoseEstimator}
We selected HybrIK \cite{li2022hybrik} for pose estimation in this study because it represents the current state-of-the-art approach. For comparison, we also conducted experiments using VIBE \cite{kocabas2019vibe}. The Mean Per Joint Position Error (MPJPE) scores for both methods are presented in Table \ref{table:InferenceTimeOfPoseEstimator}, and their respective inference times are provided in Table \ref{table:MPJPE}. From these tables, we observe that the difference in inference times between HybrIK and VIBE is relatively minor in our use case, where users upload videos and await generated instructions. Thus, our system offers flexibility: users who prioritize faster feedback can opt for a lightweight model like VIBE, while those who prefer enhanced accuracy only need to wait an additional ten seconds to benefit from HybrIK’s superior precision.

\begin{table}[htb]
\small
\centering
\begin{tabular}{c c c}
\hline 
Pose Estimator & Figure Skating (FS) & Boxing (BX) \\ \hline
HyBriK & 21sec & 35sec \\ \hline
VIBE & 10sec & 14.8sec  \\ 
    \hline
    \end{tabular}
    \caption{Inference Time Comparision of different pose estimator on Figure Skating and Boxing datasets.}
\label{table:InferenceTimeOfPoseEstimator}
\end{table}

\begin{table}[htb]
\small
\begin{tabular}{c c c c}
\hline 
Pose Estimator & 3DPW & Human3.6 & MPI-INF-3DHP \\ \hline
HyBriK & 71.6 & 47.0 & 91.0 \\ \hline
VIBE & 82.9 & 65.6 & 96.6 \\ 
    \hline
    \end{tabular}
    \caption{Comparison of HyBriK and VIBE on 3DPW, Human3.6M, and MPI-INF-3DHP Using MPJPE.}
    \label{table:MPJPE}
\end{table}

Furthermore, to evaluate the temporal smoothness of the features extracted by HyBriK and VIBE, we computed the differences per joint frame to frame in the FS and BX datasets, as shown in Table \ref{table:per-joint frame-to-frame differences} below. The metric is defined as:

\begin{equation}
\text{Smoothness} = \frac{1}{T-1} \sum_{t=1}^{T-1} \left\| J_{t+1} - J_t \right\|_2,
\end{equation}

where $J_t$ denotes the 3D joint positions at frame $t$, and $T$ is the total number of frames. A lower value indicates smoother motion over time. It can be seen that HyBriK produces significantly lower differences than VIBE, indicating smoother pose estimation on our datasets. This effectively reduces joint misalignment and skeleton drift during fast actions, making HyBriK more suitable for consistently generating accurate instructions.


\begin{table}[htb]
\small
\centering
\begin{tabular}{c c c}
\hline 
Model & Figure Skating (FS) & Boxing (BX)  \\ \hline
HyBriK & 0.0260±0.0267 & 0.0043±0.0030 \\ \hline
VIBE & 0.1340±0.1604  & 0.0396±0.0443 \\ 
    \hline
    \end{tabular}
    \caption{Per-joint frame-to-frame differences.}
    \label{table:per-joint frame-to-frame differences}
\end{table}

\par To further examine the impact of pose estimation quality on instructional output, we conducted an experiment comparing HybrIK and VIBE as the pose estimators for CoachMe. The results are presented in Table \ref{table:MetricComparisonOnPoseEstimator}.

\begin{table}
\small
\centering
\begin{tabular}{c c c c c c }
\hline 
 & B1 & B4 & RG & BS & G-eval \\ \hline
Dataset & \multicolumn{5}{c}{\textbf{Figure Skating (FS)}} \\ \hline
HyBriK & 24.7 & 2.3 & 16.9 & 26.5 & 1.73 \\ \hline
VIBE & 16.3 & 2.1 & 13.1 & 0.06 & 1.35 \\  \hline
Dataset & \multicolumn{5}{c}{\textbf{Boxing (BX) }} \\ \hline
HyBriK & 43.9 & 15.3 & 26.6 & 39.8 & 1.63 \\ \hline
VIBE & 43.1 & 13.5 & 26.0 & 36.3 & 1.65 \\ 
    \hline
    \end{tabular}
    \caption{Comparison of motion instruction generation methods on FS and BX with different pose estimator. We  evaluated the consistency between the generated instruction and the ground truth using different NLP metric and G-Eval. B1, B4, RG, and BS denote BLEU-1,
BLEU-4, ROUGE, and BertScore, respectively.}
    \label{table:MetricComparisonOnPoseEstimator}
\end{table}

\par As shown in Table \ref{table:MetricComparisonOnPoseEstimator}, instruction quality declined across metrics (BLEU, ROUGE, BERTScore) when VIBE was used instead of HybrIK. These findings suggest that in complex, high-precision domains such as professional sports, even subtle inaccuracies in pose estimation can lead to misleading or suboptimal feedback.While HybrIK consistently provides more accurate results, the performance gap—particularly in simpler cases like boxing—is relatively moderate. Reflecting this, CoachMe is designed with two pose estimation modes: a fast-response mode using VIBE for immediate feedback, and a high-precision mode using HybrIK for more accurate guidance. In this paper, we report results based on HybrIK to ensure the highest possible instruction quality and consistency across experiments.

\par While HybrIK introduces a higher computational overhead compared to lighter models, feedback from professional athletes and coaches in real-world scenarios indicates that the overall evaluation time is acceptable. We observed that the processing time did not negatively impact the usability or practicality of the system.

%% file: G-eval/consistency.tex
\begin{table*}
\small
\centering
\begin{tabular}{| p{15cm} | }
\hline 
You will be given an instruction provided by the coach. You will then be given one rephrased version of that instruction.   \\
\\
Your task is to rate the rephrased version on one metric.\\
\\
Evaluation Criteria: \\
\\
Consistency (1-5) - the factual alignment between the summary and the summarized source. A factually consistent rephrased version contains only statements that logically follow from the original instruction.  \\
\\
\\
Evaluation Steps: \\ 
\\
1. Read the coach's instruction carefully and identify the main facts and details it presents. \\
2. Read the rephrased version and compare it to the coach's original instruction. Check if the rephrased version contains any factual inaccuracies that are not supported by the original instruction. \\
3. Assign a score for consistency based on the Evaluation Criteria. \\
\\
Example: \\
Coach’s Instruction: \\
\\
\{\{Document\}\} \\
\\
Rephrased Version: \\
\\
\{\{Summary\}\} \\
\\
Evaluation Form (scores ONLY): \\
Consistency: [Insert score here] \\
\\
You only need to give a score of this example directly. \\
    \hline
    \end{tabular}
    \caption{Consistency Template}
    \label{G-eval:Consistency}
\end{table*}

%% file: G-eval/error_detection_prompt.tex
\begin{table*}
\small
\centering
\begin{tabular}{| p{15cm} | }
\hline 
You will be given an instruction provided by the coach. \\
\\
Your task is to rate the instruction on one metric.\\
\\
Evaluation Criteria: \\
\\
Error Detection (0-1) - The instruction contains any wording that can point out the error clearly. \\
0: The instruction doesn't clearly point out the error. \\
1: The instruction clearly points out the error. \\
\\
\\
Evaluation Steps: \\
\\
1. Read the coach's instruction carefully and identify the main facts and details it presents.\\
2. Check if the instruction contains any wording that can point out the error clearly. Ensure that the athlete understands exactly what the problem is after hearing the instruction. Just like these two coaching instructions: "Make sure not to position your right leg too far behind your left leg when preparing for takeoff." and "Your first jump is good. For the second jump, avoid overturning your left side before takeoff and bend your right knee more for better height." \\
3. Assign a score for Error Detection based on the Evaluation Criteria. \\
\\
Example: \\
Coach’s Instruction: \\
\\
\{\{Instruction\}\}
\\
Evaluation Form (scores ONLY):\\
Error Detection: [Insert score here]\\
\\
You only need to give a score of this example directly. \\
ONLY OUTPUT A SINGLE NUMBER (0 or 1) WITH NO ADDITIONAL TEXT. \\
    \hline
    \end{tabular}
    \caption{G-eval : Error Detection Template.}
    \label{G-eval:Error}
\end{table*}

%% file: G-eval/time_detection_prompt.tex
\begin{table*}
\small
\centering
\begin{tabular}{| p{15cm} | }
\hline 
You will be given an instruction provided by the coach.\\
\\
Your task is to rate the instruction on one metric.\\
\\
Evaluation Criteria:\\
\\
Time Detection (0-1) - The instruction contains wording that clearly includes time-related information.for example like: when take off, during the landing... \\ 
0: The instruction doesn't mention time-related information. \\
1: The instruction clearly mentions time-related information. \\
\\
\\
Evaluation Steps: \\
\\
1. Read the coach's instruction carefully and identify the main facts and details it presents. \\
2. Check if the instruction contains any wording that clearly includes time-related information. Ensure that the athlete can understand when and how to act after hearing the instruction. Just like these three coaching instructions: "Remember to keep your posture straight and vertical during the initial stages of the jump, it will help in achieving the correct form.", "Bend your hands a bit at takeoff to aid your spin." and "Stretch your right side backward and left side forward beforehand to increase power."\\
3. Assign a score for Time Detection based on the Evaluation Criteria.\\
\\
Example:\\
Coach’s Instruction: \\
\\
\{\{Instruction\}\} \\
\\
Evaluation Form (scores ONLY): \\
Time Detection: [Insert score here] \\
\\
You only need to give a score of this example directly. \\
ONLY OUTPUT A SINGLE NUMBER (0 or 1) WITH NO ADDITIONAL TEXT. \\
    \hline
    \end{tabular}
    \caption{G-eval : Time Detection Template}
    \label{G-eval:Time}
\end{table*}

%% file: G-eval/bodypart_prompt.tex
\begin{table*}
\small
\centering
\begin{tabular}{| p{15cm} | }
\hline 
You will be given an instruction provided by the coach.\\
\\
Your task is to rate the instruction on one metric.\\
\\
Evaluation Criteria:\\
\\
Body Parts Detection (0-1) - The instruction contains any wording that includes body parts clearly.\\
0: The instruction doesn't clearly point out the body parts.\\
1: The instruction clearly points out the body parts.\\
\\
\\
Evaluation Steps:\\
\\
1. Read the coach's instruction carefully and identify the main facts and details it presents. \\
2. Check if the instruction contains any wording that can point out the body parts clearly. Ensure that the athlete understands exactly which body parts to focus on after hearing the instruction. Just like these two coaching instructions: "Practice moving your left foot along the ice towards your right foot, without lifting it too soon." and "Keep your right leg from going over your left leg during takeoff for a smoother jump. Ensure your right toe lock stays firm. Manage your hands, don’t lift them too high while spinning, and make a circle with them." \\
3. Assign a score for Error Detection based on the Evaluation Criteria.\\
\\
Example:\\
Coach’s Instruction: \\
\\
\{\{Instruction\}\} \\
\\
Evaluation Form (scores ONLY): \\
Body Parts Detection: [Insert score here] \\
\\
You only need to give a score of this example directly. \\
ONLY OUTPUT A SINGLE NUMBER (0 or 1) WITH NO ADDITIONAL TEXT. \\
    \hline
    \end{tabular}
    \caption{G-eval : BodyPart Detection Template}
    \label{G-eval:BodyPart}
\end{table*}

%% file: G-eval/causation_detection_prompt.tex
\begin{table*}
\small
\centering
\begin{tabular}{| p{15cm} | }
\hline 
You will be given an instruction provided by the coach.\\
\\
Your task is to rate the instruction on one metric.\\
\\
Evaluation Criteria:\\
\\
Causation Detection (0-1) - The instruction contains any wording that can describe the logic of the motion or its cause-and-effect relationship clearly.\\
0: The instruction doesn't clearly point out the cause-and-effect relationship of the motion.\\
1: The instruction clearly points out the cause-and-effect relationship of the motion.\\
\\
\\
Evaluation Steps:\\
\\
1. Read the coach's instruction carefully and identify the main facts and details it presents.\\
2. Check if the instruction contains any wording that can describe the logic of the motion or its cause-and-effect relationship clearly. Ensure that the athlete understands exactly what the logic of the motion is after hearing the instruction. Just like these two coaching instructions: "Practice keeping your left leg closer to your right foot and lifting your left knee higher to help create more power for a higher jump because you are not fully rotating your jumps right now." and "Remember to keep your posture straight and vertical during the initial stages of the jump, it will help in achieving the correct form."\\
3. Assign a score for Error Detection based on the Evaluation Criteria.\\
\\
Example:\\
Coach’s Instruction: \\
\\
\{\{Instruction\}\} \\
\\
Evaluation Form (scores ONLY):\\
Body Parts Detection: [Insert score here]\\
\\
You only need to give a score of this example directly. \\
ONLY OUTPUT A SINGLE NUMBER (0 or 1) WITH NO ADDITIONAL TEXT. \\
    \hline
    \end{tabular}
    \caption{G-eval : Causation Detection Template}
    \label{G-eval:Causation}
\end{table*}

%% file: G-eval/method_detection_prompt.tex
\begin{table*}
\small
\centering
\begin{tabular}{| p{15cm} | }
\hline 
You will be given an instruction provided by the coach. \\
\\
Your task is to rate the instruction on one metric. \\
\\
Evaluation Criteria: \\
\\
Method Detection (0-1) - The instruction contains any wording that can clearly explains how to improve the motion. \\
\\
0: The instruction doesn't clearly explains how to improve the motion. \\
1: The instruction clearly explains how to improve the motion. \\
\\
\\
Evaluation Steps: \\
\\
1. Read the coach's instruction carefully and identify the main facts and details it presents. \\
2. Check if the instruction contains any wording that can clearly explains how to improve the motion. Ensure that the athlete understands exactly what the logic of the motion is after hearing the instruction. Just like these two coaching instructions: "Practice moving your left foot along the ice towards your right foot, without lifting it too soon." and "You need to have more speed before the jump." \\
3. Assign a score for Error Detection based on the Evaluation Criteria. \\
\\
Example: \\
Coach’s Instruction: \\
\\
\{\{Instruction\}\} \\
\\
Evaluation Form (scores ONLY):\\
Body Parts Detection: [Insert score here]\\
\\
You only need to give a score of this example directly. \\
ONLY OUTPUT A SINGLE NUMBER (0 or 1) WITH NO ADDITIONAL TEXT. \\
    \hline
    \end{tabular}
    \caption{G-eval : Method Detection Template}
    \label{G-eval:Method}
\end{table*}

%% file: G-eval/coordination_detection_prompt.tex
\begin{table*}
\small
\centering
\begin{tabular}{| p{15cm} | }
\hline 
You will be given an instruction provided by the coach. \\
\\
Your task is to rate the instruction on one metric. \\
\\
Evaluation Criteria: \\
\\
Coordination Detection (0-1) - The instruction contains any wording that can clearly explains how the body coordinates movements or how more than two body parts work together. \\
0: The instruction doesn't clearly explains how the body coordinates movements or how more than two body parts work together. \\
1: The instruction clearly explains how the body coordinates movements or how more than two body parts work together. \\
\\
\\
Evaluation Steps: \\
\\
1. Read the coach's instruction carefully and identify the main facts and details it presents. \\
2. Check if the instruction contains any wording that can clearly explains how the body coordinates movements or how more than two body parts work together. Ensure that the athlete understands exactly how to coordinate the movement after hearing the instruction. Just like these two coaching instructions: "Practice moving your left foot along the ice towards your right foot, without lifting it too soon." and "Make sure not to position your right leg too far behind your left leg when preparing for takeoff. Stretch your right side backward and left side forward beforehand to increase power. Try using more force for the takeoff and rotate your body towards the jump to raise your left leg higher." \\
3. Assign a score for Error Detection based on the Evaluation Criteria. \\
\\
Example: \\
Coach’s Instruction: \\
\\
\{\{Instruction\}\}
\\
Evaluation Form (scores ONLY): \\
Body Parts Detection: [Insert score here] \\
\\
You only need to give a score of this example directly. \\
ONLY OUTPUT A SINGLE NUMBER (0 or 1) WITH NO ADDITIONAL TEXT. \\
    \hline
    \end{tabular}
    \caption{G-eval : Coordination Detection Template}
    \label{G-eval:Coordination}
\end{table*}